\pdfoutput=1

\documentclass[11pt]{article}

\usepackage{EMNLP2022}

\usepackage{times}
\usepackage{latexsym}

\usepackage[T1]{fontenc}

\usepackage[utf8]{inputenc}

\usepackage{microtype}

\usepackage{inconsolata}

\usepackage[normalem]{ulem}
\usepackage{pifont}
\usepackage{subcaption}
\usepackage{tikzsymbols}
\newcommand{\Hquad}{\hspace{0.5em}} 

\title{\Springtree[1.75] \textsc{NarraSum}: A Large-Scale Dataset for Abstractive \\Narrative Summarization}

\author{Chao Zhao$^{1}$ \qquad Faeze Brahman$^{2,3}$ \qquad Kaiqiang Song$^{4}$ \\ \qquad {\bf Wenlin Yao}$^{4}$ \qquad {\bf Dian Yu}$^{4}$ \qquad {\bf Snigdha Chaturvedi} $^{1}$ \\
\{zhaochao, snigdha\}@cs.unc.edu \qquad faezeb@allenai.org \\ \{riversong, wenlinyao, yudian\}@global.tencent.com\\
$^{1}$ UNC Chapel Hill \Hquad $^{2}$ Allen Institute for AI \Hquad $^{3}$ University of Washington \Hquad $^{4}$ Tencent AI Lab}

\usepackage{amsmath}
\usepackage{booktabs}
\usepackage{mathtools}
\usepackage{amsfonts}
\usepackage{color}
\usepackage{tablefootnote}

\usepackage{multirow}
\usepackage{graphicx}
\usepackage{xcolor,colortbl}
\definecolor{Gray}{gray}{0.85}

\begin{document}
\maketitle
\begin{abstract}
Narrative summarization aims to produce a distilled version of a narrative to describe its most salient events and characters. Summarizing a narrative is challenging as it requires an understanding of event causality and character behaviors. To encourage research in this direction, we propose \textsc{NarraSum}, a large-scale narrative summarization dataset. 
It contains 122K narrative documents, which are collected from plot descriptions of movies and TV episodes with diverse genres, and their corresponding abstractive summaries.  
Experiments show that there is a large performance gap between humans and the state-of-the-art summarization models on \textsc{NarraSum}. We hope that this dataset will promote future research in summarization, as well as broader studies of natural language understanding and generation. The dataset is available at \url{https://github.com/zhaochaocs/narrasum}.
\end{abstract}

\section{Introduction}
\label{sec::intro}

A narrative is a story (e.g., a novel or a movie) composed of events and characters~\cite{Prince}. 
\textit{Narrative summarization} aims to produce a distilled version of a narrative, either extractively or abstractively, to contain its most salient events and major characters \cite{lehnert1981plot}.
This ability is especially crucial for the understanding of narratives, and in general, the understanding of human behaviors and beliefs \cite{piper2021narrative}. Practically, a summary of a narrative can enable a reader to quickly discern the key points, which is useful in real-world scenarios such as content recommendations and advertisements.

\begin{figure}[!t]
    \centering
    \includegraphics[width=1\linewidth]{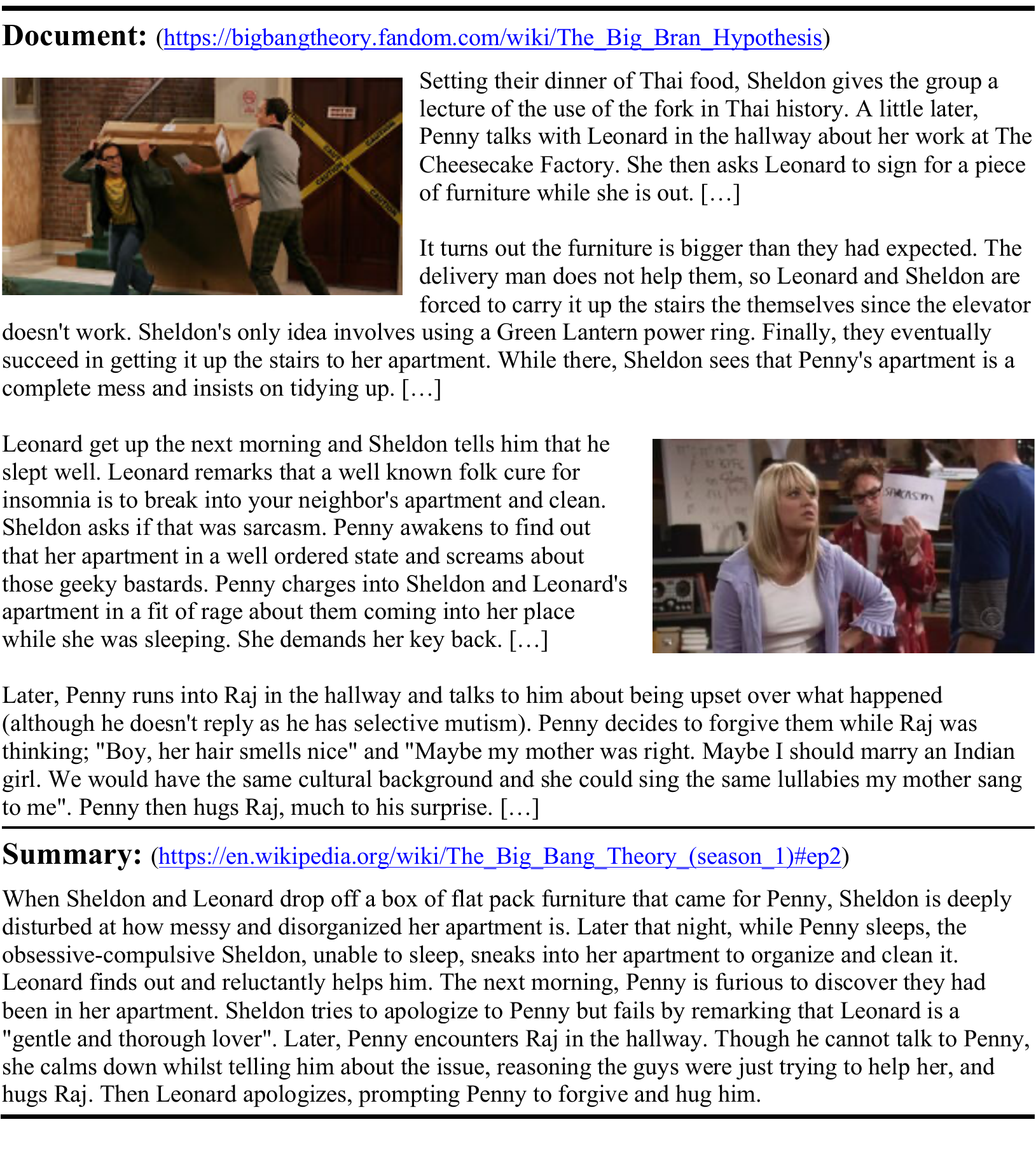} %
    \caption{Example of the narrative summarization task. The input is a narrative text (denoted by ``Document'', \textbf{pictures are not included}), and the output is a summary containing its salient events and characters. }
    \label{fig:example}
    \vspace{-10pt}
\end{figure}

While text summarization has been explored for over decades, most existing studies focus on summarizing news~\cite{linguistic2008new,nallapati-etal-2016-abstractive, narayan-etal-2018-dont} or structured documents (e.g., scientific papers \cite{gidiotis2019structured,cohan2018discourse}).
These documents have specific writing styles. For instance, news is organized such that the first few sentences convey the most important information \cite{hicks2016writing}. Scientific papers usually follow a standard structure with a few sections 
contributing the most to the summary \cite{gidiotis2020divide}. 
It has been demonstrated that many summarization models, including recent ones, heavily rely on these structural clues \cite{kedzie2018content, zhong-etal-2019-searching,zhao2022read}.
However, a typical narrative does not contain such structural cues. This suggests that a narrative summarization model has to understand the entire narrative to identify the salient events and characters. 
While some recent summarization tasks also require understanding an entire document, they focus on conversational domains such as dialogues \cite{gliwa-etal-2019-samsum}, emails \cite{zhang2021emailsum}, and meetings \cite{zhong2021qmsum}.
Narratives are different from those genres in nature and are understudied.

Understanding an entire narrative faces unique challenges. A narrative organizes the story into a sequence of events (i.e., plot) in a chronological and causal order \cite{forster1985aspects}. Events 
unfold due to the actions of characters and other event participants, or external forces in stories \cite{mani2012computational}. To identify the salient events, a model needs to understand both \textbf{plot} and \textbf{characters}. %
From the plot's perspective, the model needs to understand the causal and temporal relationships between events, as well as how the plot develops from the beginning to the end \cite{freytag1908freytag}. 
From the character's perspective, the model needs to understand the characters' profiles (e.g., personalities, roles, and interpersonal relationships), and how their desires and actions drive the story forward.%

Figure \ref{fig:example} illustrates the importance of understanding the entire narrative for summarization. 
In this example, the main event
is ``\textit{Sheldon cleans Penny's apartment and gets Leonard in trouble}'', which is included in the summary. The side event ``\textit{Penny speaks to Raj and forgives Leonard}'' is also included since it is the consequence and ending of the main event.
Whereas, \textit{``Sheldon gives a lecture of fork''} is not included as it does not impact the development of the plot. Besides the main events, the summary also explains Sheldon's motivation to clean the apartment.

A large-scale high-quality dataset is essential to promote research on this topic. Unfortunately, different from other domains, such as news and scientific papers, where the document and summary can be found from the same data source, narrative documents and their corresponding summaries are usually spread in separate sources. Previous studies collect document-summary pairs of narrative by either creating summaries manually \cite{ouyang-etal-2017-crowd} or matching titles between documents and summaries followed by a manual inspection \cite{ladhak-etal-2020-exploring,kryscinski2021booksum}, making 
it challenging to enlarge the resulting datasets.%

In this work we propose an automatic data construction framework to build a narrative summarization dataset with both large scale and high quality. %
Specifically, we first collect 
narratives from plot descriptions of movies or TV episodes through online resources. We choose the plot description because it describes the overall narrative of the movie or TV episode, including the story arcs and major characters.
This source is also widely used in narrative-related studies \cite{linebarger2009tv,bamman2013learning,papalampidi2019movie,xiong2019graph}. After data collection, we build an \textbf{align-and-verify} pipeline to automatically align plot descriptions of the same movie or TV episodes from different sources. Finally, we construct document-summary pairs by treating the long plot description as the document to be summarized and the shorter one (of the same movie or TV episode) as the corresponding summary. 
After filtering out low-quality document-summary pairs, we build \textbf{\textsc{NarraSum}}, a large-scale dataset that contains around 122K \textbf{narra}tive document-\textbf{sum}mary pairs in English. 
Our data construction framework is generic and thus can potentially be applied to other languages as well.

To gauge %
the feasibility of \textsc{NarraSum} for the narrative summarization task, we explore different characteristics of this dataset. We observe that compared with other summarization datasets, the narratives in \textsc{NarraSum} are of diverse genres, and the summaries are more abstractive and of varying lengths. 
Furthermore, rather than focusing on a particular part of the document (as in other summarization datasets), the summaries in \textsc{NarraSum} are designed to cover the entire narratives. It brings new challenges to current summarization methods. %

We investigate the performance of several strong baselines and state-of-the-art summarization models on \textsc{NarraSum}. Results show that there is a large gap between %
human and machine performance in various dimensions, demonstrating that narrative summarization is a challenging task. 

The contributions of this paper are four-fold:
\vspace{-5pt}
\begin{itemize}
\setlength\itemsep{-0.3em}
    \item We propose an automatic data construction framework to build a large-scale, high-quality narrative summarization dataset.
    \item We release the largest narrative summarization dataset to date named \textsc{NarraSum}, with detailed data analysis;
    \item We investigate the performance of recent summarization models on \textsc{NarraSum};
    \item We perform a thorough analysis of the models to point out the challenges and several promising directions.
\end{itemize}

\section{Data Construction}
We propose an automatic data construction framework to create a narrative summarization dataset. To this end, we first collect plot descriptions of movies and TV episodes from multiple resources as narratives (Section \ref{subsec: collection}). We then align plot descriptions in these resources that refer to the same movie or TV episode (Section \ref{subsec: alignment}). 
Finally, we filter the aligned data to construct high-quality document-summary pairs. (Section \ref{subsec: pairing}). We describe the details of each step as follows.

\subsection{Data Collection}
\label{subsec: collection}
We collect plot descriptions of movies and TV episodes from various movie websites and online encyclopedias such as Wikipedia,\footnote{\url{https://www.wikipedia.org/}.} Fandom,\footnote{\url{https://www.fandom.com/}.} IMDB, \footnote{\url{https://www.imdb.com/}.} TVDB, \footnote{\url{https://thetvdb.com/}.} and TMDB. \footnote{\url{https://www.themoviedb.org/}.} Note that while we use movie/TV plot descriptions as a source of narrative text, our goal is not to summarize movies and TV episodes themselves but rather to study the task of narrative summarization in a broader sense. 
Tasks of movie/TV summarization have been addressed by other datasets such as Scriptbase \cite{gorinski2015movie}, Screenplay \cite{papalampidi2020screenplay}, and SummScreen \cite{chen2022summscreen}. Those works focus more on summarizing screenplays, which describe the movements, actions, expressions, and dialogue of the characters in a specific structure and format. Compared with general narrative summarization, screenplay summarization presents a different set of challenges such as scene understanding and dialog parsing.  Plot descriptions, on the other hand, describe the movie stories from a third-person point of view and present a different set of challenges as we described in Section \ref{sec::intro}.

To collect plot descriptions, we parse web pages of movies or TV episodes based on HTML tags and use heuristics to match keywords (e.g., \textit{Synopsis}, \textit{Summary}, and \textit{Plot}) that are related to the plot. 
We then extract the texts under these sections as plot descriptions of the corresponding movies or TV episodes. 
Besides the plot descriptions, we also collect the meta information of movies or TV episodes such as their title, air date, director(s), and writer(s), which is used for data alignment.

\subsection{Data Alignment}
\label{subsec: alignment}

After data collection, we align the web pages that are from different websites but refer to the same movie or TV episode. It is a challenging task due to the ambiguity in natural language%
. For example, a single movie may have different surface forms of the title (e.g., \textit{Avengers 4} and \textit{Avengers: Endgame}), while those with the same title may refer to different movies (e.g., \textit{Bad Company} may refer to fourteen movies.) Similar ambiguity issues arise when aligning air dates or names of crew members. %
Also, meta-information might be missing or incorrect due to the editing or parsing mistakes of web pages. To address these challenges, we propose an \textbf{align-and-verify} pipeline. It first aligns movie or TV episodes via fuzzy meta-information matching, which encourages high recall. Then, we use a verifier with high precision to re-check the aligned pairs and filter out the pairs with low confidence. We describe the details of this pipeline as follows.

During the \textbf{alignment} stage, we apply several heuristics for fuzzy meta-information matching. To align movies, we
first normalize movie titles by removing non-alphanumeric characters, stopwords, and subtitles. We then collect the movie pairs where the Levenshtein distance between the normalized titles is less than a threshold.\footnote{We set the threshold to be $0.2\times l$, where $l$ is the maximum length of the two titles. All thresholds in this section were chosen by experimenting with different values and manually analyzing the quality of a subset of the data.} Besides the title match, we also require the two movies to have the same air date or a partial overlap on directors or writers when such information is available. 
The ambiguity in titles of TV episodes is more severe than that of movies. To align TV episodes, we apply similar heuristics and further require the two episodes to belong to the same TV show.

During the \textbf{verification} stage, we improve the precision of alignment by comparing %
the aligned plot descriptions. 
Specifically, we train a classifier to take as input the concatenation of two plot descriptions to predict if they should be aligned. To train such a classifier, we first build a dataset with balanced positive aligned pairs and negative pairs. The positive pairs are a subset of heuristically aligned pairs where there is an link in one web page (e.g., ``External links'' in Wikipedia) pointing to the web page of the same movie or TV episode in the other website. Such links are edited by humans and are commonly used in entity linking \cite{shen2014entity}. 
Negative pairs are randomly sampled from different movies of the same movie series or different episodes of the same TV show. Negative pairs sampled by this strategy usually share a similar set of characters and background setting, preventing the model from relying on surface-level cues to solve the task. %

Based on the data sampling method, we collected a large-scale balanced dataset with $60$K positive pairs and $60$K negative pairs. We then split the dataset into train/validation/test subsets with the ratio of $80$\%/$10$\%/$10$\%.
We train a RoBERTa-base~\cite{Liu2019RoBERTaAR} classifier on this dataset and it achieves an accuracy of $97.13$\% on the test set, indicating that this model can serve as a reliable verifier to improve the precision of data alignment. 
We employ this classifier to further verify the heuristically aligned plot descriptions and filter out those where the predicted log-odds is smaller than 1. %
Finally, we obtain $2.6$ million aligned plot description pairs. %

\subsection{Document-Summary Pairing}
\label{subsec: pairing}
After obtaining the aligned plot description pairs, we regard the longer plot description as the document and the shorter one as the corresponding summary. However, not all pairs are of good quality for summarization. We identify three major issues compromising the quality and remove the relatively low-quality pairs from the final dataset.

First, the summary may contain hallucinated content that might not be included in the document. Similar to \cite{ladhak-etal-2020-exploring},
we observe that hallucination is less common in plot description pairs with a noticeable difference in length. 
We therefore require the length of the summary to be shorter than half of the document to be summarized. We also calculate the semantic matching score between a summary and a document, and then remove the pairs with low scores. We adopt two scores. The first is the Rouge-1 Precision between the summary and the document. The second is the entailment probability between the summary and the document obtained from DocNLI \cite{yin-etal-2021-docnli}, a document-level NLI model. We add up the two scores, rank the instances accordingly, and remove the $3\%$ document-summary pairs with the lowest score.

Second, sometimes the content in the shorter plot description is directly copied from the longer plot description. To create an abstractive summarization dataset, we use ROUGE-2 Precision \cite{lin2004rouge} between the document and the summary to reflect whether the content of the summary is copied from the document, and remove the pairs where the ROUGE-2 Precision is larger than 0.5.

Third, a plot description may only describe part of the entire narrative such as a trailer but does not necessarily summarize the narrative. 
To filter out these cases, we set the minimum length of documents and summaries to make sure that they contain enough information. \footnote{For movies, we set the minimum length of documents and summaries as 200 and 100. For TV episodes, we set the minimum length as 100 and 50.}
We also extract oracle extractive summaries from the original document using the method proposed by~\citet{liu2019text}.  We remove the instances where less than 30\% content of the oracle extractive summaries are from either the first half or the second half of the document.

After applying these filtering strategies, we obtain the final version of \textsc{NarraSum}. It contains 122K aligned document-summary pairs, which is a high-quality subset (3.8\%) of the original aligned pairs. We split the dataset into training (90\%), validation (5\%), and testing (5\%) sets at the title level in order to avoid data leakage and undesirable overlap between training and validation or test sets.

\section{Data Analysis}
This section provides basic statistics of \textsc{NarraSum}. We then analyze the dataset in terms of the distribution of salient information and abstractiveness of summaries. Finally, we conduct a human assessment to evaluate the quality of \textsc{NarraSum}.

\subsection{Data Statistics}

\begin{table}[t!]
		\centering
		\small
		\setlength{\tabcolsep}{0.4em} %
		\begin{tabular}{l|ccccc}
\toprule \textbf{Datasets} & \textbf{Domain} & \textbf{Size}  & \textbf{L-doc} & \textbf{L-sum} & \textbf{Ratio} \\
\midrule CNNDM & News & $312$K  & 781 & 56 & $13.9$  \\
XSum & News & 227K  & 431 & 20 & $21.5$  \\\hline
 arXiv & Sci-Paper & 215K  & 4,938 & 220 & $22.4$  \\
 PubMed & Sci-Paper & 133K  & 3,016 & 203 & $14.9$  \\\hline
 NovelChap & Novel & 8K & 5,165 & 372 & $13.9$  \\
 BookSum & Novel & 12K  & 5,102 & 505 & $10.1$  \\
\hline \bf \textsc{NarraSum} & Movie/TV & 122K & 786 &  147 & 5.3 \\

\bottomrule
\end{tabular}

		\caption{\label{tab::stat} Comparison between \textsc{NarraSum} and other datasets accroding to the domain, size, document length, summary length, and compression ratio.} %
		\vspace{-6pt}
\end{table}

\begin{figure}[!t]
    \centering
    
    \includegraphics[width=1\linewidth]{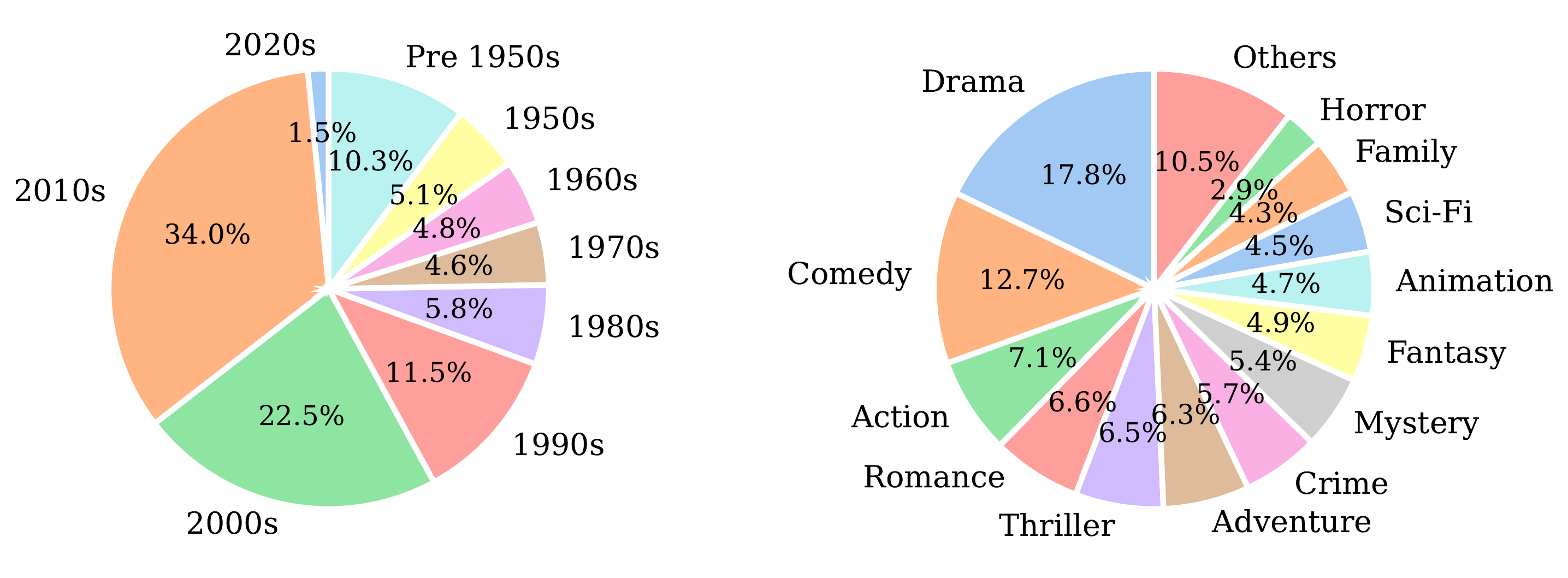} 
  
    \caption{Distribution of production years and genres in \textsc{NarraSum}.}
    \label{fig:year-genre}
\end{figure}
 
We compare \textsc{NarraSum} with six datasets from different domains such as news, scientific papers, and narratives. These include CNN DailyMail (CNNDM) \cite{see-etal-2017-get}, XSum \cite{narayan2018don}, ArXiv \cite{cohan2018discourse}, PubMed \cite{cohan2018discourse}, NovelChapter \cite{ladhak-etal-2020-exploring}, and BookSum \cite{kryscinski2021booksum}. The comparison of statistics is shown in Table \ref{tab::stat}.

\begin{table}[t]
		\centering
		\small
		\setlength{\tabcolsep}{0.55em} %
		\begin{tabular}{l|cccc}
\toprule \multirow{2}{*}{ \textbf{Datasets} } & \multicolumn{4}{c}{ \textbf{\% of novel n-grams in summary} } \\
& 1-grams & 2-grams & 3-grams & 4-grams \\
\midrule CNN/DM & $17.00$ & $53.91$ & $71.98$ & $80.29$ \\
XSum & $35.76$ & $83.45$ & $95.50$ & $98.49$ \\
Pubmed & $18.53$ & $48.23$ & $68.28$ & $78.39$ \\\hline
\bf \textsc{NarraSum} & $47.78$ & $81.86$ & $94.96$ & $98.00$ \\
\bottomrule
\end{tabular}
		\caption{\label{tab::ngram} Comparison of novel n-grams between \textsc{NarraSum} and other summarization datasets.}
\end{table}

\textsc{NarraSum} contains 122K instances from 22.8K unique movies and 28.5K unique TV episodes, which is ten times larger than the previous largest narrative summarization dataset.
We provide the distribution of production years and genres of these movies or TV series in Figure \ref{fig:year-genre}, which illustrates that \textsc{NarraSum} spans a wide time period and contains a broad range of genres.
The average length of documents and summaries are 785.97 and 147.06 tokens, and the average compression ratio is %
5.34. Most of the documents in \textsc{NarraSum} are longer than 512 tokens, which is the maximum input length of many pre-trained language models. However, the average length of documents in \textsc{NarraSum} is still shorter than that of a typical novel chapter ($\sim$5K).
This requires the models to process long, but not prohibitively long, inputs while exposing them to the challenges of narrative summarization.

\subsection{Summary Characteristics}
\begin{figure}[!t]
    \centering
    \subfloat{\includegraphics[width=1\linewidth]{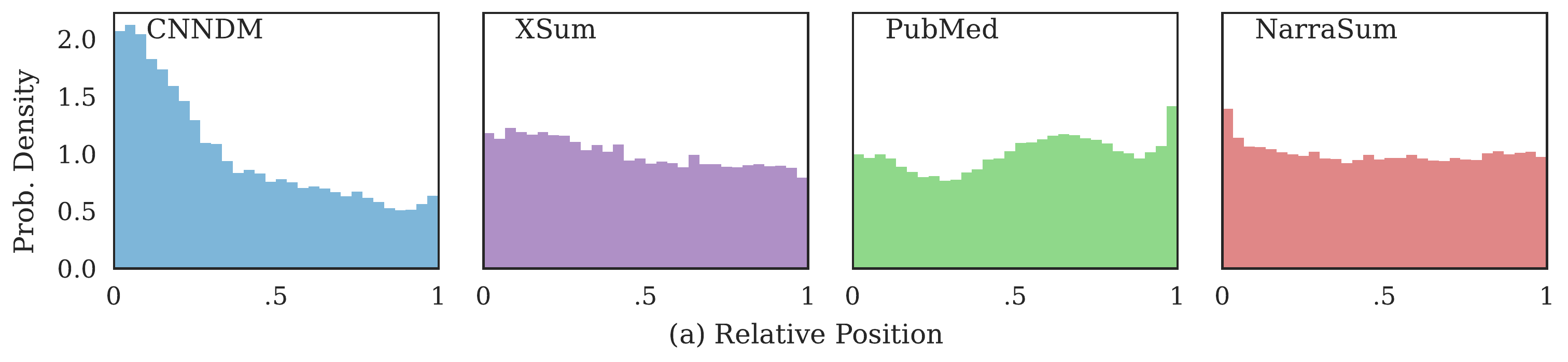} \label{fig:a}} \\\vspace{-6pt}
    \subfloat{\includegraphics[width=1\linewidth]{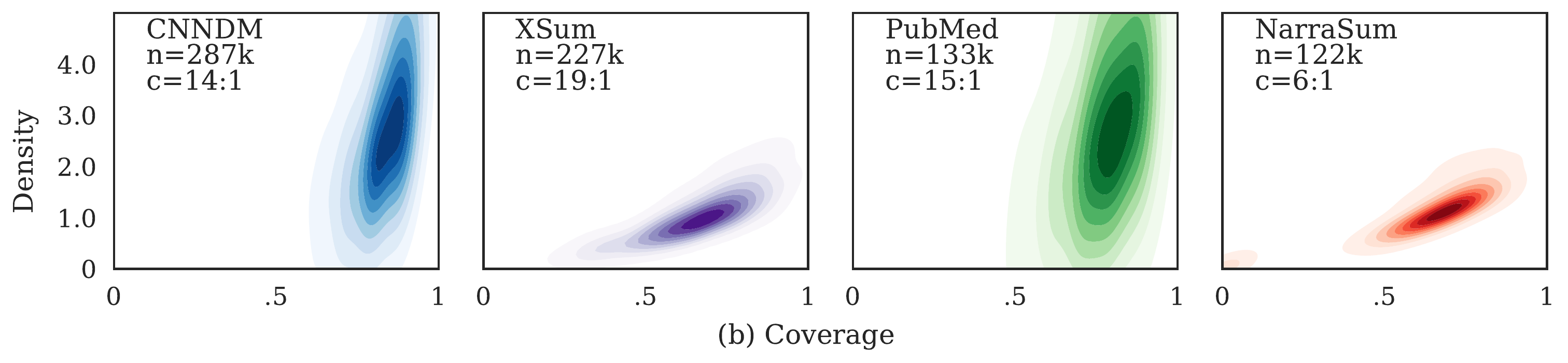} \label{fig:b}}

    \caption{The upper figures show the relative positions of bi-grams of the gold summary in the document. The summary content of \textsc{NarraSum} is more uniformly distributed over the entire document. The lower figures show the Coverage-Density plots. Compared with CNNDM and PubMed, the summary abstractiveness of \textsc{NarraSum} is more close to XSum.}
    \label{fig:cov-den}
\end{figure}

Different from news articles, salient information in a narrative spreads across the entire text. To verify whether \textsc{NarraSum}'s summaries have this property%
, we first check the \textbf{distribution of the salient information} in the documents. Similar to \citet{kim2019abstractive}, we use bi-grams of summary text to represent the salient content of the narrative and then obtain their normalized positions in the documents. Figure \ref{fig:cov-den}(a) shows the probability density distribution of the positions of the salient information. We compare the distribution of \textsc{NarraSum} with CNNDM, XSum, and PubMed. Figure \ref{fig:cov-den}(a) indicates that while the salient information of CNNDM and PubMed are concentrated at certain parts of the document, the salient information of \textsc{NarraSum} is more uniformly distributed over the entire document.
It supports our claim that the summarization of \textsc{NarraSum} requires an understanding of the entire document. %
There is no lead bias in XSum because the first sentence of the document is removed and is regarded as the summary. It further demonstrates that the first sentence of a news document is enough to summarize the entire document. The section-wise bias in scientific papers is discussed by \citet{gidiotis2020divide}.%

Next, we measure the \textbf{abstractiveness of summaries} in \textsc{NarraSum}. To this end, we calculate the Coverage and Density of each summary as suggested by \citet{grusky2018newsroom}. Lower Coverage and Density scores indicate that the summary is more abstractive. The distribution is shown in Figure \ref{fig:cov-den}(b). The comparison shows that the summaries of \textsc{NarraSum} are more abstractive than CNNDM and PubMed while being similar to XSum, the most abstractive dataset for news summarization.

We also report the percentage of novel n-grams that are included in the summary but not in the document. A higher percentage of novel n-grams implies a more abstractive summary.
As shown in Table \ref{tab::ngram}, the percentage of novel n-grams in \textsc{NarraSum} is higher than CNNDM and PubMed, and is similar to XSum. This is in line with our observation from the Coverage-Density plot (Figure~\ref{fig:cov-den}(b)). The difference is that XSum is a news summarization dataset with short summaries (one sentence). \textsc{NarraSum} is a narrative summarization dataset, where the summaries are of varying length.

\subsection{Quality Assessment}
We further conduct a human evaluation to better assess the quality of the \textsc{NarraSum}. 
We randomly select $100$ instances from the test set. For each instance, we ask three workers on Amazon Mechanical Turk to evaluate the summary in terms of \textit{faithfulness} and \textit{informativeness}. For faithfulness, we show annotators each summary sentence and ask them to evaluate how much of the information in this summary sentence is presented in the document. This is a precision-oriented measure and is commonly used for summary evaluation \cite{lu2020multi}. 
For informativeness, we ask annotators to first identify the most salient events and major characters from the document and then evaluate how much of that is covered by the summary. %
This is a recall-oriented measure. 
Both human evaluations are collected on a Likert scale of 1-5 (1 means ``none'', and 5 means ``almost all''). 

To control the annotation quality, we require human judges to be in the United States, and have more than 1,000 HITs approved with an approval rate higher than 98\%. We randomly check the annotation results and block the human judges who continually provide low-quality annotations.
Human judges were paid a wage rate of \$12 per hour, which is higher than the local minimum wage rate.

\begin{figure}[t]
    \centering
    \includegraphics[width=1\linewidth]{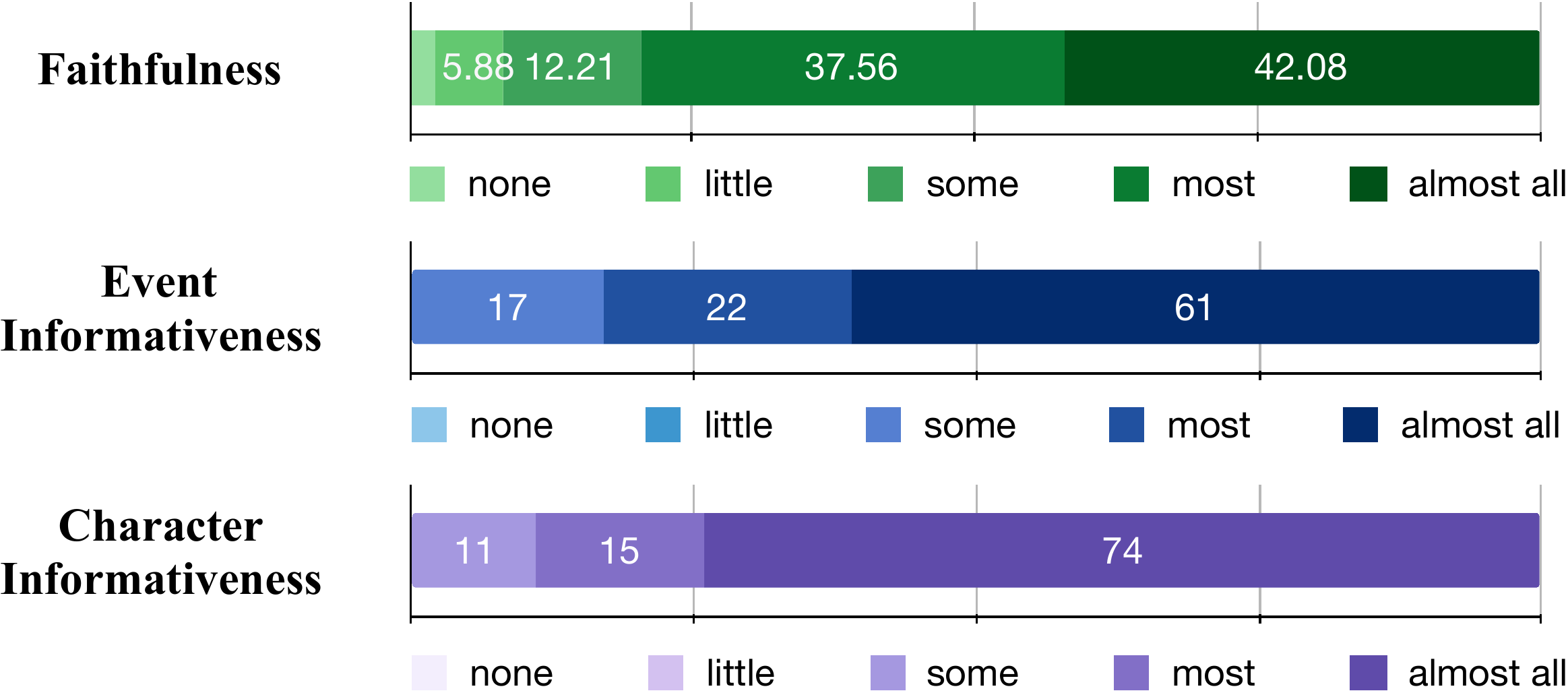} %
    \caption{Human assessment results of the quality of \textsc{NarraSum}.}
    \label{fig:data_eval}
\end{figure}

Figure \ref{fig:data_eval} shows the distributions of human evaluation results. It shows that 80\%
of content in the summary is faithful to the document.
For informativeness, 83\% and 89\% of summaries cover most of the salient events and characters, respectively. It demonstrates that \textsc{NarraSum} is of high quality in both faithfulness and informativeness, and can foster further research on narrative summarization. 

\section{Baseline Models}
\label{sec::method}
We investigate the performance of several baselines and state-of-the-art neural summarization models on \textsc{NarraSum}. We include both extractive and abstractive models. For extractive models, we use the following methods:

\noindent\textbf{\textsc{Random}} selects $n$ sentences from the document randomly.

\noindent\textbf{\textsc{Lead}} selects the top-$n$ sentences from the document to compose the summary. This is a strong baseline for news summarization.

\noindent\textbf{\textsc{TextRank}} \cite{mihalcea2004textrank} is a graph-based extractive summarization model based on PageRank \cite{brin1998anatomy} in a graph representation of sentences.

\noindent\textbf{\textsc{LexRank}} \cite{erkan2004lexrank} is another graph-based extractive summarization model based on eigenvector centrality .

\noindent\textbf{\textsc{HSG}} \cite{wang2020heterogeneous} is a heterogeneous graph-based neural extractive summarization model that uses word co-occurrence  to enhance sentence contextual representation.%

\noindent\textbf{\textsc{PreSumm}} \cite{liu2019text} relies on a pre-trained language model to enhance the sentence representation during text encoding and extractive summarization. We choose \textbf{\textsc{BERT}}~\cite{devlin-etal-2019-bert}, \textbf{\textsc{RoBERTa}}~\cite{Liu2019RoBERTaAR}, and \textbf{\textsc{Longformer}}~\cite{Beltagy2020LongformerTL} as the pre-trained models. BERT and RoBERTa limit the input length to be shorter than 512 tokens, while Longformer can accept up to 4,096 tokens.  %

For abstractive models, we use the following pre-trained sequence-to-sequence models: \textbf{\textsc{BART}}~\cite{lewis-etal-2020-bart}, \textbf{\textsc{T5}}~\cite{2020t5}, \textbf{\textsc{Pegasus}}~\cite{zhang2019pegasus}, and \textbf{\textsc{LED}}~\cite{Beltagy2020LongformerTL}. The input length of the first three models is limited to 512 (base version) or 1,024 (large version). \textsc{LED} uses Longformer as the encoder and therefore can accept up to 4,096 tokens as input.

\section{Experiments}

\begin{table}[t]

		\small
	\begin{center}
\renewcommand{\arraystretch}{1.2}
\begin{tabular}{l cccc}
\toprule
    Model
    & \textbf{\textsc{R-1}} & \textbf{\textsc{R-2}} & \textbf{\textsc{R-L}} &\textbf{\textsc{SC}} \\
\midrule
\rowcolor{Gray!93}\textit{\textbf{Extractive}}& & & &\\
            \textsc{Rand} & 33.94 & 5.38 & 29.80 & - \\
\textsc{Lead} & 35.11 & 6.71 & 30.82 & - \\
\textsc{LexRank} & 34.22 & 5.78 & 29.70 & -\\
\textsc{TextRank} & 34.95 & 6.18 & 30.28 & -\\ \hline
\textsc{HSG} & 36.94 & 7.54 & 32.35 & - \\
\textsc{Bert-Base} & 36.34 & 7.29 & 31.71 & -\\
\textsc{Roberta-Base} & 36.47 & 7.31 & 31.80 & -\\
\textsc{LFormer-Base} & \textbf{37.54}* & \textbf{7.83}* & \textbf{32.69}* & - \\\hline
\textsc{Oracle} & 42.42 & 11.44 & 36.65 & -\\
			\midrule
\rowcolor{Gray!93}\textit{\textbf{Abstractive}}& & & &\\
\textsc{Bart-Base} & 35.81 & 7.49 & 31.72 & 65.19 \\
\textsc{T5-Base} & 36.37 & 7.42 & 32.17 & 76.38 \\
\textsc{LED-Base} & 37.32 & 8.14 & 33.05 & 62.63 \\\hline
\textsc{Bart-Large} & 36.80 & 8.20 & 32.62 & \textbf{77.41}* \\
\textsc{T5-Large}  & 37.67 & 8.11 & \textbf{33.40} & 74.14 \\
\textsc{Pegasus-Large} & 36.97 & 7.93 & 32.64 &  75.23 \\
\textsc{LED-Large} & \textbf{37.71} & \textbf{8.87}* & 33.34 & 66.91 \\
 \bottomrule[1pt]

		\end{tabular}
		
	\end{center}
	\caption{\label{tab:results} Summarization results evaluated on test set of \textsc{NarraSum} over ROUGE 1 (R-1), ROUGE 2 (R-2), ROUGE L (R-L), and SummaC (SC). SC is only used to evaluate abstractive summaries as extractive summaries are faithful by design. We highlight the best scores separately for extractive and abstrative systems. * indicates a statistically significant difference compared with the second best score (bootstrap resampling, $p<0.05$ \cite{koehn2006manual}). }
	\vspace{-1em}
\end{table}

\subsection{Settings}

We conduct experiments with models described in Section \ref{sec::method} to evaluate their performances on \textsc{NarraSum}. For extractive models, we follow the hyper-parameters of the original implementations. 
For abstractive models, we implement them using the Transformer library~\cite{Wolf2019HuggingFacesTS}. We fine-tune each model on the training set of \textsc{NarraSum} with AdamW optimizer~\cite{loshchilov2018decoupled} and batch size of 64. 
We conduct a simple hyper-parameter search for the learning rate from $\{3e^{-4}, 1e^{-4}, 3e^{-5}\}$ based on the validation loss.
We also adopt early stopping based on the validation loss to avoid overfitting. During inference, we use beam search with beam-size 5. %
Our model was trained on a single Quadro RTX 5000 GPU in up to 34 hours, depending on the model size. 

\noindent \textbf{Evaluation}. \space We evaluate the generated summaries using ROUGE $F_1$ score.\footnote{\url{https://github.com/google-research/google-research/tree/master/rouge}.} 
We further include SummaC \cite{laban2022summac}, an automatic measure for summary faithfulness. It achieves state-of-the-art on the benchmark of summary inconsistency detection, and is feasible to be applied to long input and output. %

\subsection{Automatic Results}
Table \ref{tab:results} shows the results on \textsc{NarraSum} using extractive and abstractive summarization approaches. 

\noindent \textbf{Extractive Models}. The supervised extractive methods outperform the unsupervised extractive methods (the first four models) on all measures by a large margin, indicating that \textsc{NarraSum} can provide a strong supervision signal for identifying the salient information and creating the summary accordingly.
PreSumm-BERT or PreSumm-Roberta models underperform HSG
because these models have a maximum input length of 512 tokens whereas HSG can accept inputs with arbitrary length.
Longformer achieves the best performance on extractive summarization by combining the advantage of pre-training and long document processing. However, there is still a large gap between Longformer's performance and the oracle upper-bound, indicating the challenges in narrative summarization.

\noindent \textbf{Abstractive Models}. Among these models, no particular model consistently outperforms others on all subsets. Larger models consistently outperform smaller models, which is inline with previous research.
T5 outperforms BART on most Rouge scores, as they adopt summarization-specific pre-training objectives. LED outperforms other models on Rouge due to its ability to encode longer documents. This is consistent with the result of extractive summarization. However, LED performs worst on SummaC-based faithfulness evaluation. This indicates that though the model can process longer documents, understanding and faithfully summarizing lengthy texts is still challenging. %

\noindent \textbf{Compression Degree}. 
To better understand the models’ capability under different compression degrees, we split the test set into three similar-sized subsets based on the compression ratio of the summary. We then re-evaluate models on each subset separately. We provide details of data split and model performance in Appendix \ref{app::split}. Results show that it is more challenging to create a short summary than a long one. Other observations on the entire test set still hold across subsets with different levels of compression.

\subsection{Human Evaluation}

We further conduct a human evaluation on Amazon Mechanical Turk to better understand the models' behaviors and the challenges of this task. We randomly sample 100 instances from the test set and then evaluate the outputs of the best two systems (T5-Large and LED-Large) based on the following four dimensions.
\begin{itemize}
\setlength\itemsep{-0.1em}
    \item  Fluency: whether or not the summary is grammatically correct and free of repetition;
    \item  Faithfulness: whether or not the summary is faithful to the original document; 
    \item  Coherence: whether or not the plot of the narrative summary is logically coherent;
    \item  Informativeness: whether or not the summary reflects the salient events and characters in the original document;
\end{itemize}

\begin{table}[t]
		\centering
		\small
		\begin{tabular}{l|cccc}
\toprule
Model & T5-Large & LED-Large \\
\midrule
Fluency & 4.19 & 4.11 \\
Faithfulness & 3.34 & 3.23 \\
Coherence & 2.87 & 3.06 \\
Informativeness & 2.44 & 2.67 \\
\bottomrule
\end{tabular}

		\caption{\label{tab::human_eval} Human evaluation of the generated summaries. }
\end{table}

For each instance, we show annotators the original document and the generated summaries. We ask annotators to rate summaries using a 5-point Likert scale and report the average score over all instances. As shown in Table \ref{tab::human_eval}, while the pre-trained abstractive models are good at Fluency, they still struggle with other dimensions such as Faithfulness, Coherence, and Informativeness. It further indicates that narrative summarization is a challenging task for current models. In general, the summaries created by T5 are more fluent and faithful, while those created by LED are more coherent and informative. In appendix \ref{app::case}, we provide examples of generated summaries by various systems.

\section{Analysis}
We perform a series of analyses about the summary position and character consistency. For a fair comparison among models, we only choose test instances where the length of the document is shorter than the maximum input length of these models (1,024 tokens). 

\subsection{Analysis of Summary Position}

A good narrative summary should preserve the original narrative structure that contains a start, middle, and ending of the narrative.
To investigate this, we adopt the method in \citet{kim2019abstractive} to analyze the normalized position of summary bi-grams in the document, where 0 and 1 represent the start and ending of the document, respectively. 

\begin{figure}[!t]
    \centering
    
    \includegraphics[width=1\linewidth]{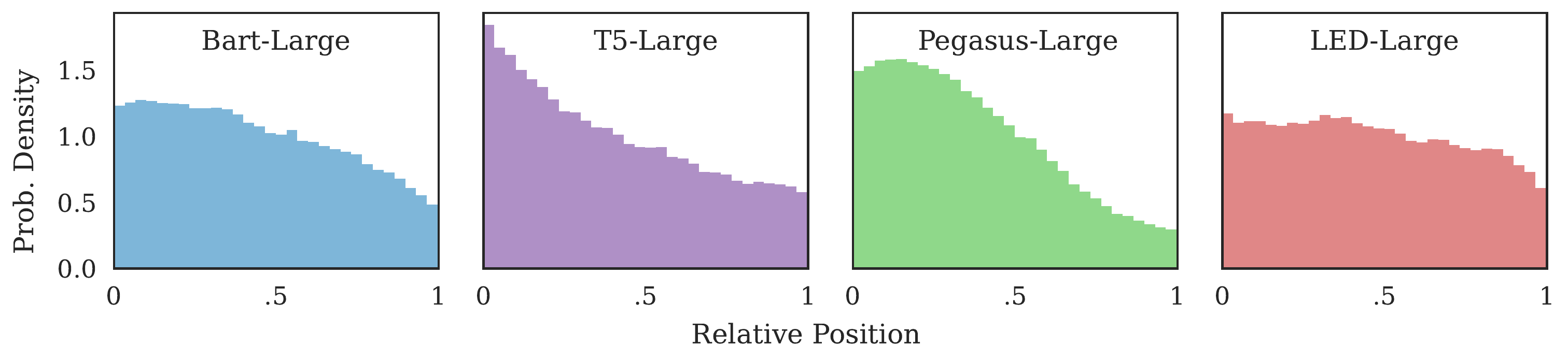} 
  
    \caption{The relative positions of bi-grams of the predicted summaries in the document. }
    \label{fig:summary_dist}
\end{figure}

Figure \ref{fig:summary_dist} shows that while the relative position of n-grams in gold summary is more close to uniformly distributed (Figure \ref{fig:cov-den}(a)), the generated summaries are still biased towards the beginning of the original document. It indicates that current models have difficulty understanding the entire documents and preserving the narrative structures.

\begin{table*}[t]
		\centering
  \small
		\setlength{\tabcolsep}{1.5em} %

\begin{tabular}{l|cccccc}
\toprule
{Evaluated $\rightarrow$} & MCTest & MovieQA & LiSCU & CBT & QuAIL & Reddit \\
Trained $\downarrow$ & Accuracy & Accuracy& Accuracy& Accuracy& Accuracy& Rouge-1 \\
\midrule
NovelChapter & 69.66 & 54.60 & 25.81 & 79.90 & 56.95 & 28.91\\
BookSum & 70.50 & 55.21 & 26.75 & 80.24 & 56.33 & 26.08\\\hline
\bf \textsc{NarraSum} & 71.83 & 56.64 & 26.85 & 80.66 & 57.37 & 32.80\\
\bottomrule
\end{tabular}

		\caption{\label{tab::zero-shot} Zero-shot performance (Accuracy or Rouge-1) of the model trained on NarraSum and those on other summarization datasets. }
\end{table*}

\subsection{Character-Wise Analysis}
Characters are essential for narratives. Since characters are not considered in Rouge scores, here we propose to measure character consistency by examining whether the major characters in the document are also mentioned in the summary.
We assume that major characters appear more frequently in the narrative text. By comparing the distance between the frequency distributions of characters from the document and the summary, we can understand how well the summary includes the major characters of the document. 

To this end, we first identify characters from the narrative. We run a coreference resolution model to extract clusters of entity mentions, and we only keep person entities to obtain clusters of characters.\footnote{We use CoreNLP for coreference resolution and named entity recognition.} We regard each cluster size as the frequency of the corresponding character and then normalize it as a probability. We measure the character inconsistency as the cross-entropy (CE) between the two frequency distributions of characters.
A higher CE implies a higher character inconsistency. 

\begin{figure}[!t]
    \centering
    
    \includegraphics[width=1\linewidth]{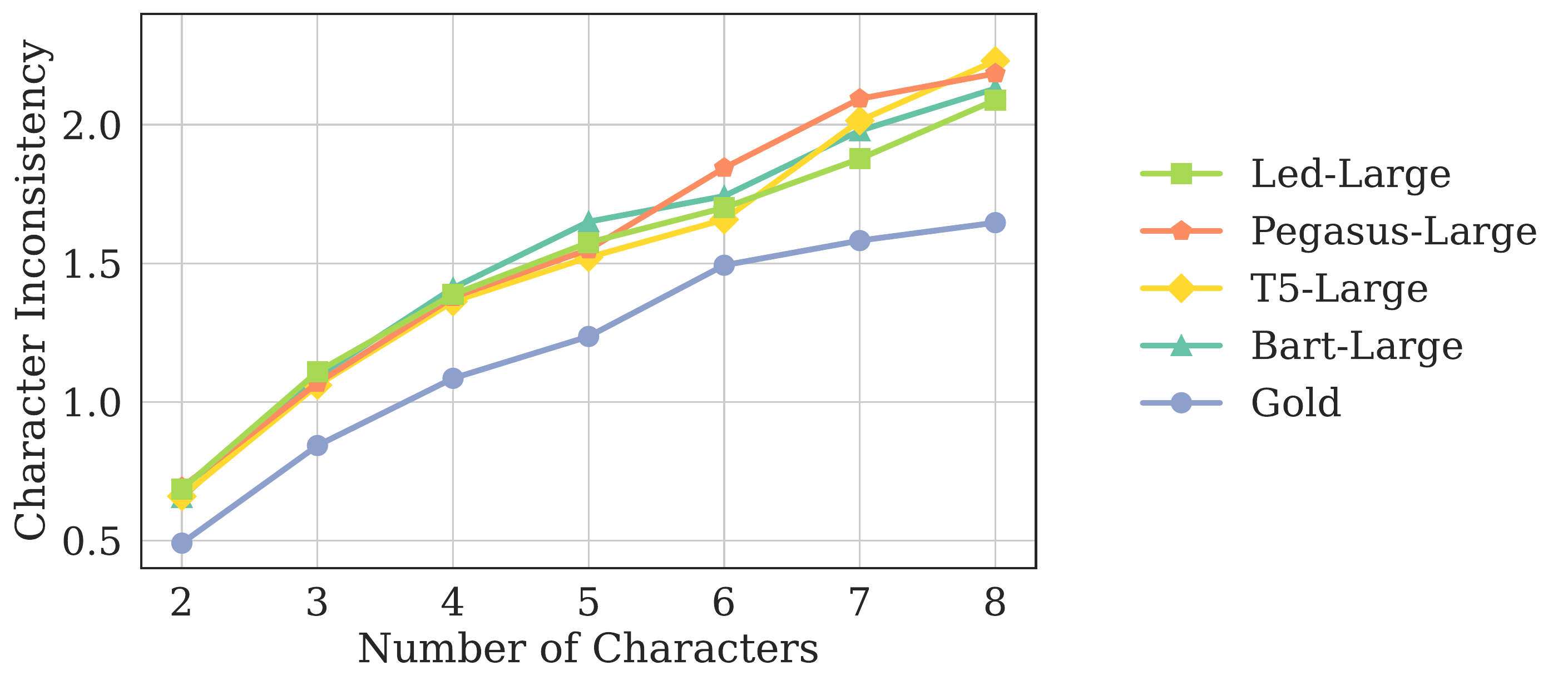} 
    \caption{Character inconsistency between documents and summaries w.r.t. the number of characters in the document.}
    \label{fig:char_num}
\end{figure}

In Figure \ref{fig:char_num}, we group the test instances of \textsc{NarraSum} based on the number of distinct characters, and show the cross-entropy of the gold summary and the generated summaries.
Compared with the gold summaries, the generated summaries are less consistent with the document at the character level. 
In general, the difference of cross-entropy between gold summary and generated summaries increases as the number of characters increases, indicating that it is harder for the summarizer to keep the character-level consistency when the document describes more characters.

\section{Application to Other Tasks}

\begin{table}[t]
		\centering
  
\small
		\setlength{\tabcolsep}{0.75em} %
		\renewcommand{\arraystretch}{1.2}
\begin{tabular}{l ccc}
\toprule
    Model
    & \textbf{\textsc{R-1}} & \textbf{\textsc{R-2}} & \textbf{\textsc{R-L}} \\
\midrule
Novel Chapter & 32.56 & 6.83 & 16.25\\
\quad w/ \textsc{NarraSum} pretraining & 32.88 & 6.80 & 16.19 \\ \hline
BookSum-Paragraph & 21.17 & 4.35 & 16.78\\
\quad w/ \textsc{NarraSum} pretraining & 21.83 & 4.86 & 17.13\\
\bottomrule
\end{tabular}
		\caption{\label{tab::results_narrative} Model performance on Novel Chapter and BookSum-Paragraph with and without pretraining on \textsc{NarraSum}. \vspace{-6pt}}
\end{table}

Besides presenting \textsc{NarraSum} as a benchmark for narrative summarization, we further explore the broader benefits of this dataset to narrative-related tasks. We first investigate whether pre-training on \textsc{NarraSum} can improve performance on other narrative summarization tasks. To this end, we first pre-train a BART-Large model on \textsc{NarraSum} and then finetune it on Novel Chapter and BookSum-Paragraph. We compare with the finetuned models without pre-training on \textsc{NarraSum}. As shown in Table \ref{tab::results_narrative}, pre-training on \textsc{NarraSum} can improve model performance on both datasets, indicating that \textsc{NarraSum} is beneficial to other narrative summarization tasks.

We then investigate if \textsc{NarraSum} can help the model learn general knowledge of narrative understanding and summarization.
For this, we first pre-train a BART-Large model on \textsc{NarraSum} and then apply it to several downstream tasks in a zero-shot manner.
We choose five tasks that are designed for narrative understanding, i.e., MCTest \cite{richardson2013mctest}, MovieQA \cite{tapaswi2016movieqa} , LiSCU \cite{brahman2021let}, CBT \cite{hill2016goldilocks}, and QuAIL \cite{rogers2020getting}, and one task for narrative summarization, i.e., Reddit TIFU \cite{kim2019abstractive}.
For each task, we provide the corresponding task description, method, and evaluation measure in Appendix \ref{app::zero-shot}.

We use models trained on the summarization task to solve these tasks in a zero-shot manner. In other words, we do not use any training data from these tasks. For discriminative tasks, we first convert the (question, answer) pair into a statement using a T5 model \cite{chen-etal-2021-nli-models}, and then evaluate the probability of generating each statement conditioned on the document \cite{zhao2022learning}. We choose the candidate with the highest generation probability as the predicted answer. Models are evaluated using Accuracy. 
For the summarization task, we directly apply the trained model to create the summary. Models are evaluated using the Rouge-1 F measure.

We compare the model pre-trained on \textsc{NarraSum} with those pre-trained on other narrative summarization datasets such as Novel Chapter and BookSum. As shown in Table \ref{tab::zero-shot}, the model pre-trained on \textsc{NarraSum} achieves better performance on all narrative-related downstream tasks compared with those pre-trained on other datasets. It indicates that \textsc{NarraSum} contains high-quality knowledge about narrative understanding and summarization, which can be beneficial to general narrative-related tasks as well.

\section{Conclusion}
We present \textsc{NarraSum}, a large-scale narrative summarization dataset that contains plot descriptions of movies and TV episodes and the corresponding summaries. Narratives in \textsc{NarraSum} are of diverse genres, and the summaries are highly abstractive and of varying lengths. Summarizing the narratives in \textsc{NarraSum} requires narrative-level understanding, which poses new challenges to current summarization methods. 
Experiments show that current models struggle with creating high-quality narrative summaries. 
We hope that \textsc{NarraSum} will promote future research in text summarization, as well as broader NLP studies such as machine reading comprehension, narrative understanding, and creative writing.

\section*{Acknowledgements}
This work was supported in part by NSF grant DRL-2112635.
We thank anonymous reviewers for their thoughtful and constructive reviews.

\section*{Limitations}
One limitation of \textsc{NarraSum}, similar to other automatically constructed datasets, is that we cannot guarantee the entire faithfulness of the summary to the document. To alleviate this issue, we first collect a large-scale dataset and then apply strict rules to select a high-quality subset. The human evaluation and the comparison with other datasets demonstrate that it is worth the trade-off. Another limitation is that \textsc{NarraSum} does not cover all narrative types such as books, scripts, and personal stories. For those purposes, we suggest readers explore other summarization datasets \cite{gorinski2015movie,ouyang-etal-2017-crowd,kim2019abstractive,ladhak-etal-2020-exploring,papalampidi2020screenplay,kryscinski2021booksum,chen2022summscreen}.

\section*{Broader Impact}
Besides the contribution to the research field of text summarization, this dataset may spark interest in a broader NLP community. For example, in machine reading comprehension, our paired plot descriptions with low lexical overlap can improve the model's capacity of complex reasoning and understanding \cite{saha2018duorc}. In narrative understanding, a summary of the narrative can help identify the salient event \cite{zhang2021salience} as well as the causal, temporal, and hierarchical relationships of events \cite{hidey-mckeown-2016-identifying, yao2020weakly}. In creative writing and storytelling, this dataset can support the research of expanding a short story outline to a more detailed story \cite{ammanabrolu2020story}. 

We collect and use the publicly available resources for research purposes only, which belong to fair use. 
This dataset should not be deployed in the real world as anything other than a research prototype, especially commercially.

There is the possibility of (potentially harmful) social biases that can exist in the movies or TV episodes and therefore be introduced in the dataset. While such biases have a limited impact on summarization systems (e.g., introducing harmful biases to the summary when there are no such biases in the document), we suggest the users evaluate the biases and their impacts on their downstream tasks such as creative writing and storytelling, and to make modifications to either the dataset or their models accordingly to avoid such biases.

\bibliography{anthology,custom}

\begin{thebibliography}{65}
\expandafter\ifx\csname natexlab\endcsname\relax\def\natexlab#1{#1}\fi

\bibitem[{Ammanabrolu et~al.(2020)Ammanabrolu, Tien, Cheung, Luo, Ma, Martin,
  and Riedl}]{ammanabrolu2020story}
Prithviraj Ammanabrolu, Ethan Tien, Wesley Cheung, Zhaochen Luo, William Ma,
  Lara~J. Martin, and Mark~O. Riedl. 2020.
\newblock \href {https://aaai.org/ojs/index.php/AAAI/article/view/6232} {Story
  realization: Expanding plot events into sentences}.
\newblock In \emph{The Thirty-Fourth {AAAI} Conference on Artificial
  Intelligence, {AAAI} 2020, The Thirty-Second Innovative Applications of
  Artificial Intelligence Conference, {IAAI} 2020, The Tenth {AAAI} Symposium
  on Educational Advances in Artificial Intelligence, {EAAI} 2020, New York,
  NY, USA, February 7-12, 2020}, pages 7375--7382. {AAAI} Press.

\bibitem[{Bamman et~al.(2013)Bamman, O{'}Connor, and
  Smith}]{bamman2013learning}
David Bamman, Brendan O{'}Connor, and Noah~A. Smith. 2013.
\newblock \href {https://aclanthology.org/P13-1035} {Learning latent personas
  of film characters}.
\newblock In \emph{Proceedings of the 51st Annual Meeting of the Association
  for Computational Linguistics (Volume 1: Long Papers)}, pages 352--361,
  Sofia, Bulgaria. Association for Computational Linguistics.

\bibitem[{Beltagy et~al.(2020)Beltagy, Peters, and
  Cohan}]{Beltagy2020LongformerTL}
Iz~Beltagy, Matthew~E. Peters, and Arman Cohan. 2020.
\newblock Longformer: The long-document transformer.
\newblock \emph{ArXiv}, abs/2004.05150.

\bibitem[{Brahman et~al.(2021)Brahman, Huang, Tafjord, Zhao, Sachan, and
  Chaturvedi}]{brahman2021let}
Faeze Brahman, Meng Huang, Oyvind Tafjord, Chao Zhao, Mrinmaya Sachan, and
  Snigdha Chaturvedi. 2021.
\newblock \href {https://doi.org/10.18653/v1/2021.findings-emnlp.150} {{``}let
  your characters tell their story{''}: A dataset for character-centric
  narrative understanding}.
\newblock In \emph{Findings of the Association for Computational Linguistics:
  EMNLP 2021}, pages 1734--1752, Punta Cana, Dominican Republic. Association
  for Computational Linguistics.

\bibitem[{Brin and Page(1998)}]{brin1998anatomy}
Sergey Brin and Lawrence Page. 1998.
\newblock The anatomy of a large-scale hypertextual web search engine.
\newblock \emph{Computer networks and ISDN systems}, 30(1-7):107--117.

\bibitem[{Chen et~al.(2021)Chen, Choi, and Durrett}]{chen-etal-2021-nli-models}
Jifan Chen, Eunsol Choi, and Greg Durrett. 2021.
\newblock \href {https://doi.org/10.18653/v1/2021.findings-emnlp.324} {Can
  {NLI} models verify {QA} systems{'} predictions?}
\newblock In \emph{Findings of the Association for Computational Linguistics:
  EMNLP 2021}, pages 3841--3854, Punta Cana, Dominican Republic. Association
  for Computational Linguistics.

\bibitem[{Chen et~al.(2022)Chen, Chu, Wiseman, and Gimpel}]{chen2022summscreen}
Mingda Chen, Zewei Chu, Sam Wiseman, and Kevin Gimpel. 2022.
\newblock \href {https://doi.org/10.18653/v1/2022.acl-long.589}
  {{S}umm{S}creen: A dataset for abstractive screenplay summarization}.
\newblock In \emph{Proceedings of the 60th Annual Meeting of the Association
  for Computational Linguistics (Volume 1: Long Papers)}, pages 8602--8615,
  Dublin, Ireland. Association for Computational Linguistics.

\bibitem[{Cohan et~al.(2018)Cohan, Dernoncourt, Kim, Bui, Kim, Chang, and
  Goharian}]{cohan2018discourse}
Arman Cohan, Franck Dernoncourt, Doo~Soon Kim, Trung Bui, Seokhwan Kim, Walter
  Chang, and Nazli Goharian. 2018.
\newblock \href {https://doi.org/10.18653/v1/N18-2097} {A discourse-aware
  attention model for abstractive summarization of long documents}.
\newblock In \emph{Proceedings of the 2018 Conference of the North {A}merican
  Chapter of the Association for Computational Linguistics: Human Language
  Technologies, Volume 2 (Short Papers)}, pages 615--621, New Orleans,
  Louisiana. Association for Computational Linguistics.

\bibitem[{Consortium and Company(2008)}]{linguistic2008new}
Linguistic~Data Consortium and New York~Times Company. 2008.
\newblock \href {https://books.google.com/books?id=D4F2AQAACAAJ} {\emph{The New
  York Times Annotated Corpus}}.
\newblock LDC corpora. Linguistic Data Consortium.

\bibitem[{Devlin et~al.(2019)Devlin, Chang, Lee, and
  Toutanova}]{devlin-etal-2019-bert}
Jacob Devlin, Ming-Wei Chang, Kenton Lee, and Kristina Toutanova. 2019.
\newblock \href {https://doi.org/10.18653/v1/N19-1423} {{BERT}: Pre-training of
  deep bidirectional transformers for language understanding}.
\newblock In \emph{Proceedings of the 2019 Conference of the North {A}merican
  Chapter of the Association for Computational Linguistics: Human Language
  Technologies, Volume 1 (Long and Short Papers)}, pages 4171--4186,
  Minneapolis, Minnesota. Association for Computational Linguistics.

\bibitem[{Erkan and Radev(2004)}]{erkan2004lexrank}
G{\"u}nes Erkan and Dragomir~R Radev. 2004.
\newblock Lexrank: Graph-based lexical centrality as salience in text
  summarization.
\newblock \emph{Journal of artificial intelligence research}, 22:457--479.

\bibitem[{Fan et~al.(2018)Fan, Grangier, and Auli}]{fan2018controllable}
Angela Fan, David Grangier, and Michael Auli. 2018.
\newblock \href {https://doi.org/10.18653/v1/W18-2706} {Controllable
  abstractive summarization}.
\newblock In \emph{Proceedings of the 2nd Workshop on Neural Machine
  Translation and Generation}, pages 45--54, Melbourne, Australia. Association
  for Computational Linguistics.

\bibitem[{Forster(1985)}]{forster1985aspects}
Edward~Morgan Forster. 1985.
\newblock \emph{Aspects of the Novel}, volume~19.
\newblock Houghton Mifflin Harcourt.

\bibitem[{Freytag(1908)}]{freytag1908freytag}
Gustav Freytag. 1908.
\newblock \emph{Freytag's technique of the drama: an exposition of dramatic
  composition and art}.
\newblock Scott, Foresman and Company.

\bibitem[{Gidiotis and Tsoumakas(2019)}]{gidiotis2019structured}
Alexios Gidiotis and Grigorios Tsoumakas. 2019.
\newblock Structured summarization of academic publications.
\newblock In \emph{Joint European Conference on Machine Learning and Knowledge
  Discovery in Databases}, pages 636--645. Springer.

\bibitem[{Gidiotis and Tsoumakas(2020)}]{gidiotis2020divide}
Alexios Gidiotis and Grigorios Tsoumakas. 2020.
\newblock A divide-and-conquer approach to the summarization of long documents.
\newblock \emph{IEEE/ACM Transactions on Audio, Speech, and Language
  Processing}, 28:3029--3040.

\bibitem[{Gliwa et~al.(2019)Gliwa, Mochol, Biesek, and
  Wawer}]{gliwa-etal-2019-samsum}
Bogdan Gliwa, Iwona Mochol, Maciej Biesek, and Aleksander Wawer. 2019.
\newblock \href {https://doi.org/10.18653/v1/D19-5409} {{SAMS}um corpus: A
  human-annotated dialogue dataset for abstractive summarization}.
\newblock In \emph{Proceedings of the 2nd Workshop on New Frontiers in
  Summarization}, pages 70--79, Hong Kong, China. Association for Computational
  Linguistics.

\bibitem[{Gorinski and Lapata(2015)}]{gorinski2015movie}
Philip~John Gorinski and Mirella Lapata. 2015.
\newblock \href {https://doi.org/10.3115/v1/N15-1113} {Movie script
  summarization as graph-based scene extraction}.
\newblock In \emph{Proceedings of the 2015 Conference of the North {A}merican
  Chapter of the Association for Computational Linguistics: Human Language
  Technologies}, pages 1066--1076, Denver, Colorado. Association for
  Computational Linguistics.

\bibitem[{Grusky et~al.(2018)Grusky, Naaman, and Artzi}]{grusky2018newsroom}
Max Grusky, Mor Naaman, and Yoav Artzi. 2018.
\newblock \href {https://doi.org/10.18653/v1/N18-1065} {{N}ewsroom: A dataset
  of 1.3 million summaries with diverse extractive strategies}.
\newblock In \emph{Proceedings of the 2018 Conference of the North {A}merican
  Chapter of the Association for Computational Linguistics: Human Language
  Technologies, Volume 1 (Long Papers)}, pages 708--719, New Orleans,
  Louisiana. Association for Computational Linguistics.

\bibitem[{Hicks et~al.(2016)Hicks, Sally, Gilbert, Holmes, and
  Bentley}]{hicks2016writing}
Wynford Hicks, Adams Sally, Harriett Gilbert, Tim Holmes, and Jane Bentley.
  2016.
\newblock \emph{Writing for journalists}.
\newblock Routledge.

\bibitem[{Hidey and McKeown(2016)}]{hidey-mckeown-2016-identifying}
Christopher Hidey and Kathy McKeown. 2016.
\newblock \href {https://doi.org/10.18653/v1/P16-1135} {Identifying causal
  relations using parallel {W}ikipedia articles}.
\newblock In \emph{Proceedings of the 54th Annual Meeting of the Association
  for Computational Linguistics (Volume 1: Long Papers)}, pages 1424--1433,
  Berlin, Germany. Association for Computational Linguistics.

\bibitem[{Hill et~al.(2016)Hill, Bordes, Chopra, and
  Weston}]{hill2016goldilocks}
Felix Hill, Antoine Bordes, Sumit Chopra, and Jason Weston. 2016.
\newblock \href {http://arxiv.org/abs/1511.02301} {The goldilocks principle:
  Reading children's books with explicit memory representations}.
\newblock In \emph{4th International Conference on Learning Representations,
  {ICLR} 2016, San Juan, Puerto Rico, May 2-4, 2016, Conference Track
  Proceedings}.

\bibitem[{Kedzie et~al.(2018)Kedzie, McKeown, and
  Daum{\'e}~III}]{kedzie2018content}
Chris Kedzie, Kathleen McKeown, and Hal Daum{\'e}~III. 2018.
\newblock \href {https://doi.org/10.18653/v1/D18-1208} {Content selection in
  deep learning models of summarization}.
\newblock In \emph{Proceedings of the 2018 Conference on Empirical Methods in
  Natural Language Processing}, pages 1818--1828, Brussels, Belgium.
  Association for Computational Linguistics.

\bibitem[{Kim et~al.(2019)Kim, Kim, and Kim}]{kim2019abstractive}
Byeongchang Kim, Hyunwoo Kim, and Gunhee Kim. 2019.
\newblock \href {https://doi.org/10.18653/v1/N19-1260} {Abstractive
  summarization of {R}eddit posts with multi-level memory networks}.
\newblock In \emph{Proceedings of the 2019 Conference of the North {A}merican
  Chapter of the Association for Computational Linguistics: Human Language
  Technologies, Volume 1 (Long and Short Papers)}, pages 2519--2531,
  Minneapolis, Minnesota. Association for Computational Linguistics.

\bibitem[{Koehn and Monz(2006)}]{koehn2006manual}
Philipp Koehn and Christof Monz. 2006.
\newblock \href {https://aclanthology.org/W06-3114} {Manual and automatic
  evaluation of machine translation between {E}uropean languages}.
\newblock In \emph{Proceedings on the Workshop on Statistical Machine
  Translation}, pages 102--121, New York City. Association for Computational
  Linguistics.

\bibitem[{Kry{\'s}ci{\'n}ski et~al.(2021)Kry{\'s}ci{\'n}ski, Rajani, Agarwal,
  Xiong, and Radev}]{kryscinski2021booksum}
Wojciech Kry{\'s}ci{\'n}ski, Nazneen Rajani, Divyansh Agarwal, Caiming Xiong,
  and Dragomir Radev. 2021.
\newblock \href {https://arxiv.org/abs/2105.08209} {Booksum: A collection of
  datasets for long-form narrative summarization}.
\newblock \emph{ArXiv preprint}, abs/2105.08209.

\bibitem[{Laban et~al.(2022)Laban, Schnabel, Bennett, and
  Hearst}]{laban2022summac}
Philippe Laban, Tobias Schnabel, Paul~N. Bennett, and Marti~A. Hearst. 2022.
\newblock \href {https://doi.org/10.1162/tacl_a_00453} {{S}umma{C}: Re-visiting
  {NLI}-based models for inconsistency detection in summarization}.
\newblock \emph{Transactions of the Association for Computational Linguistics},
  10:163--177.

\bibitem[{Ladhak et~al.(2020)Ladhak, Li, Al-Onaizan, and
  McKeown}]{ladhak-etal-2020-exploring}
Faisal Ladhak, Bryan Li, Yaser Al-Onaizan, and Kathleen McKeown. 2020.
\newblock \href {https://doi.org/10.18653/v1/2020.acl-main.453} {Exploring
  content selection in summarization of novel chapters}.
\newblock In \emph{Proceedings of the 58th Annual Meeting of the Association
  for Computational Linguistics}, pages 5043--5054, Online. Association for
  Computational Linguistics.

\bibitem[{Lehnert(1981)}]{lehnert1981plot}
Wendy~G Lehnert. 1981.
\newblock Plot units and narrative summarization.
\newblock \emph{Cognitive science}, 5(4):293--331.

\bibitem[{Lewis et~al.(2020)Lewis, Liu, Goyal, Ghazvininejad, Mohamed, Levy,
  Stoyanov, and Zettlemoyer}]{lewis-etal-2020-bart}
Mike Lewis, Yinhan Liu, Naman Goyal, Marjan Ghazvininejad, Abdelrahman Mohamed,
  Omer Levy, Veselin Stoyanov, and Luke Zettlemoyer. 2020.
\newblock \href {https://doi.org/10.18653/v1/2020.acl-main.703} {{BART}:
  Denoising sequence-to-sequence pre-training for natural language generation,
  translation, and comprehension}.
\newblock In \emph{Proceedings of the 58th Annual Meeting of the Association
  for Computational Linguistics}, pages 7871--7880, Online. Association for
  Computational Linguistics.

\bibitem[{Lin(2004)}]{lin2004rouge}
Chin-Yew Lin. 2004.
\newblock \href {https://aclanthology.org/W04-1013} {{ROUGE}: A package for
  automatic evaluation of summaries}.
\newblock In \emph{Text Summarization Branches Out}, pages 74--81, Barcelona,
  Spain. Association for Computational Linguistics.

\bibitem[{Linebarger and Piotrowski(2009)}]{linebarger2009tv}
Deborah~L Linebarger and Jessica~Taylor Piotrowski. 2009.
\newblock Tv as storyteller: How exposure to television narratives impacts
  at-risk preschoolers' story knowledge and narrative skills.
\newblock \emph{British journal of developmental psychology}, 27(1):47--69.

\bibitem[{Liu and Lapata(2019)}]{liu2019text}
Yang Liu and Mirella Lapata. 2019.
\newblock \href {https://doi.org/10.18653/v1/D19-1387} {Text summarization with
  pretrained encoders}.
\newblock In \emph{Proceedings of the 2019 Conference on Empirical Methods in
  Natural Language Processing and the 9th International Joint Conference on
  Natural Language Processing (EMNLP-IJCNLP)}, pages 3730--3740, Hong Kong,
  China. Association for Computational Linguistics.

\bibitem[{Liu et~al.(2019)Liu, Ott, Goyal, Du, Joshi, Chen, Levy, Lewis,
  Zettlemoyer, and Stoyanov}]{Liu2019RoBERTaAR}
Yinhan Liu, Myle Ott, Naman Goyal, Jingfei Du, Mandar Joshi, Danqi Chen, Omer
  Levy, Mike Lewis, Luke Zettlemoyer, and Veselin Stoyanov. 2019.
\newblock Roberta: A robustly optimized bert pretraining approach.
\newblock \emph{ArXiv}, abs/1907.11692.

\bibitem[{Loshchilov and Hutter(2019)}]{loshchilov2018decoupled}
Ilya Loshchilov and Frank Hutter. 2019.
\newblock \href {https://openreview.net/forum?id=Bkg6RiCqY7} {Decoupled weight
  decay regularization}.
\newblock In \emph{7th International Conference on Learning Representations,
  {ICLR} 2019, New Orleans, LA, USA, May 6-9, 2019}. OpenReview.net.

\bibitem[{Lu et~al.(2020)Lu, Dong, and Charlin}]{lu2020multi}
Yao Lu, Yue Dong, and Laurent Charlin. 2020.
\newblock \href {https://doi.org/10.18653/v1/2020.emnlp-main.648}
  {Multi-{XS}cience: A large-scale dataset for extreme multi-document
  summarization of scientific articles}.
\newblock In \emph{Proceedings of the 2020 Conference on Empirical Methods in
  Natural Language Processing (EMNLP)}, pages 8068--8074, Online. Association
  for Computational Linguistics.

\bibitem[{Mani(2012)}]{mani2012computational}
Inderjeet Mani. 2012.
\newblock Computational modeling of narrative.
\newblock \emph{Synthesis Lectures on Human Language Technologies},
  5(3):1--142.

\bibitem[{Mihalcea and Tarau(2004)}]{mihalcea2004textrank}
Rada Mihalcea and Paul Tarau. 2004.
\newblock \href {https://aclanthology.org/W04-3252} {{T}ext{R}ank: Bringing
  order into text}.
\newblock In \emph{Proceedings of the 2004 Conference on Empirical Methods in
  Natural Language Processing}, pages 404--411, Barcelona, Spain. Association
  for Computational Linguistics.

\bibitem[{Nallapati et~al.(2016)Nallapati, Zhou, dos Santos, Gul{\c{c}}ehre,
  and Xiang}]{nallapati-etal-2016-abstractive}
Ramesh Nallapati, Bowen Zhou, Cicero dos Santos, {\c{C}}a{\u{g}}lar
  Gul{\c{c}}ehre, and Bing Xiang. 2016.
\newblock \href {https://doi.org/10.18653/v1/K16-1028} {Abstractive text
  summarization using sequence-to-sequence {RNN}s and beyond}.
\newblock In \emph{Proceedings of the 20th {SIGNLL} Conference on Computational
  Natural Language Learning}, pages 280--290, Berlin, Germany. Association for
  Computational Linguistics.

\bibitem[{Narayan et~al.(2018{\natexlab{a}})Narayan, Cohen, and
  Lapata}]{narayan-etal-2018-dont}
Shashi Narayan, Shay~B. Cohen, and Mirella Lapata. 2018{\natexlab{a}}.
\newblock \href {https://doi.org/10.18653/v1/D18-1206} {Don{'}t give me the
  details, just the summary! topic-aware convolutional neural networks for
  extreme summarization}.
\newblock In \emph{Proceedings of the 2018 Conference on Empirical Methods in
  Natural Language Processing}, pages 1797--1807, Brussels, Belgium.
  Association for Computational Linguistics.

\bibitem[{Narayan et~al.(2018{\natexlab{b}})Narayan, Cohen, and
  Lapata}]{narayan2018don}
Shashi Narayan, Shay~B. Cohen, and Mirella Lapata. 2018{\natexlab{b}}.
\newblock \href {https://doi.org/10.18653/v1/D18-1206} {Don{'}t give me the
  details, just the summary! topic-aware convolutional neural networks for
  extreme summarization}.
\newblock In \emph{Proceedings of the 2018 Conference on Empirical Methods in
  Natural Language Processing}, pages 1797--1807, Brussels, Belgium.
  Association for Computational Linguistics.

\bibitem[{Ouyang et~al.(2017)Ouyang, Chang, and
  McKeown}]{ouyang-etal-2017-crowd}
Jessica Ouyang, Serina Chang, and Kathy McKeown. 2017.
\newblock \href {https://aclanthology.org/E17-2008} {Crowd-sourced iterative
  annotation for narrative summarization corpora}.
\newblock In \emph{Proceedings of the 15th Conference of the {E}uropean Chapter
  of the Association for Computational Linguistics: Volume 2, Short Papers},
  pages 46--51, Valencia, Spain. Association for Computational Linguistics.

\bibitem[{Papalampidi et~al.(2020)Papalampidi, Keller, Frermann, and
  Lapata}]{papalampidi2020screenplay}
Pinelopi Papalampidi, Frank Keller, Lea Frermann, and Mirella Lapata. 2020.
\newblock \href {https://doi.org/10.18653/v1/2020.acl-main.174} {Screenplay
  summarization using latent narrative structure}.
\newblock In \emph{Proceedings of the 58th Annual Meeting of the Association
  for Computational Linguistics}, pages 1920--1933, Online. Association for
  Computational Linguistics.

\bibitem[{Papalampidi et~al.(2019)Papalampidi, Keller, and
  Lapata}]{papalampidi2019movie}
Pinelopi Papalampidi, Frank Keller, and Mirella Lapata. 2019.
\newblock \href {https://doi.org/10.18653/v1/D19-1180} {Movie plot analysis via
  turning point identification}.
\newblock In \emph{Proceedings of the 2019 Conference on Empirical Methods in
  Natural Language Processing and the 9th International Joint Conference on
  Natural Language Processing (EMNLP-IJCNLP)}, pages 1707--1717, Hong Kong,
  China. Association for Computational Linguistics.

\bibitem[{Piper et~al.(2021)Piper, So, and Bamman}]{piper2021narrative}
Andrew Piper, Richard~Jean So, and David Bamman. 2021.
\newblock \href {https://doi.org/10.18653/v1/2021.emnlp-main.26} {Narrative
  theory for computational narrative understanding}.
\newblock In \emph{Proceedings of the 2021 Conference on Empirical Methods in
  Natural Language Processing}, pages 298--311, Online and Punta Cana,
  Dominican Republic. Association for Computational Linguistics.

\bibitem[{Prince(1973)}]{Prince}
Gerald Prince. 1973.
\newblock \emph{A Grammar of Stories: An Introduction}.

\bibitem[{Raffel et~al.(2020)Raffel, Shazeer, Roberts, Lee, Narang, Matena,
  Zhou, Li, and Liu}]{2020t5}
Colin Raffel, Noam Shazeer, Adam Roberts, Katherine Lee, Sharan Narang, Michael
  Matena, Yanqi Zhou, Wei Li, and Peter~J. Liu. 2020.
\newblock \href {http://jmlr.org/papers/v21/20-074.html} {Exploring the limits
  of transfer learning with a unified text-to-text transformer}.
\newblock \emph{Journal of Machine Learning Research}, 21(140):1--67.

\bibitem[{Richardson et~al.(2013)Richardson, Burges, and
  Renshaw}]{richardson2013mctest}
Matthew Richardson, Christopher~J.C. Burges, and Erin Renshaw. 2013.
\newblock \href {https://aclanthology.org/D13-1020} {{MCT}est: A challenge
  dataset for the open-domain machine comprehension of text}.
\newblock In \emph{Proceedings of the 2013 Conference on Empirical Methods in
  Natural Language Processing}, pages 193--203, Seattle, Washington, USA.
  Association for Computational Linguistics.

\bibitem[{Rogers et~al.(2020)Rogers, Kovaleva, Downey, and
  Rumshisky}]{rogers2020getting}
Anna Rogers, Olga Kovaleva, Matthew Downey, and Anna Rumshisky. 2020.
\newblock \href {https://aaai.org/ojs/index.php/AAAI/article/view/6398}
  {Getting closer to {AI} complete question answering: {A} set of prerequisite
  real tasks}.
\newblock In \emph{The Thirty-Fourth {AAAI} Conference on Artificial
  Intelligence, {AAAI} 2020, The Thirty-Second Innovative Applications of
  Artificial Intelligence Conference, {IAAI} 2020, The Tenth {AAAI} Symposium
  on Educational Advances in Artificial Intelligence, {EAAI} 2020, New York,
  NY, USA, February 7-12, 2020}, pages 8722--8731. {AAAI} Press.

\bibitem[{Saha et~al.(2018)Saha, Aralikatte, Khapra, and
  Sankaranarayanan}]{saha2018duorc}
Amrita Saha, Rahul Aralikatte, Mitesh~M. Khapra, and Karthik Sankaranarayanan.
  2018.
\newblock \href {https://doi.org/10.18653/v1/P18-1156} {{D}uo{RC}: Towards
  complex language understanding with paraphrased reading comprehension}.
\newblock In \emph{Proceedings of the 56th Annual Meeting of the Association
  for Computational Linguistics (Volume 1: Long Papers)}, pages 1683--1693,
  Melbourne, Australia. Association for Computational Linguistics.

\bibitem[{See et~al.(2017)See, Liu, and Manning}]{see-etal-2017-get}
Abigail See, Peter~J. Liu, and Christopher~D. Manning. 2017.
\newblock \href {https://doi.org/10.18653/v1/P17-1099} {Get to the point:
  Summarization with pointer-generator networks}.
\newblock In \emph{Proceedings of the 55th Annual Meeting of the Association
  for Computational Linguistics (Volume 1: Long Papers)}, pages 1073--1083,
  Vancouver, Canada. Association for Computational Linguistics.

\bibitem[{Shen et~al.(2014)Shen, Wang, and Han}]{shen2014entity}
Wei Shen, Jianyong Wang, and Jiawei Han. 2014.
\newblock Entity linking with a knowledge base: Issues, techniques, and
  solutions.
\newblock \emph{IEEE Transactions on Knowledge and Data Engineering},
  27(2):443--460.

\bibitem[{Tapaswi et~al.(2016)Tapaswi, Zhu, Stiefelhagen, Torralba, Urtasun,
  and Fidler}]{tapaswi2016movieqa}
Makarand Tapaswi, Yukun Zhu, Rainer Stiefelhagen, Antonio Torralba, Raquel
  Urtasun, and Sanja Fidler. 2016.
\newblock \href {https://doi.org/10.1109/CVPR.2016.501} {Movieqa: Understanding
  stories in movies through question-answering}.
\newblock In \emph{2016 {IEEE} Conference on Computer Vision and Pattern
  Recognition, {CVPR} 2016, Las Vegas, NV, USA, June 27-30, 2016}, pages
  4631--4640. {IEEE} Computer Society.

\bibitem[{Wang et~al.(2020)Wang, Liu, Zheng, Qiu, and
  Huang}]{wang2020heterogeneous}
Danqing Wang, Pengfei Liu, Yining Zheng, Xipeng Qiu, and Xuanjing Huang. 2020.
\newblock \href {https://doi.org/10.18653/v1/2020.acl-main.553} {Heterogeneous
  graph neural networks for extractive document summarization}.
\newblock In \emph{Proceedings of the 58th Annual Meeting of the Association
  for Computational Linguistics}, pages 6209--6219, Online. Association for
  Computational Linguistics.

\bibitem[{Wolf et~al.(2020)Wolf, Debut, Sanh, Chaumond, Delangue, Moi, Cistac,
  Rault, Louf, Funtowicz, Davison, Shleifer, von Platen, Ma, Jernite, Plu, Xu,
  Le~Scao, Gugger, Drame, Lhoest, and Rush}]{Wolf2019HuggingFacesTS}
Thomas Wolf, Lysandre Debut, Victor Sanh, Julien Chaumond, Clement Delangue,
  Anthony Moi, Pierric Cistac, Tim Rault, Remi Louf, Morgan Funtowicz, Joe
  Davison, Sam Shleifer, Patrick von Platen, Clara Ma, Yacine Jernite, Julien
  Plu, Canwen Xu, Teven Le~Scao, Sylvain Gugger, Mariama Drame, Quentin Lhoest,
  and Alexander Rush. 2020.
\newblock \href {https://doi.org/10.18653/v1/2020.emnlp-demos.6} {Transformers:
  State-of-the-art natural language processing}.
\newblock In \emph{Proceedings of the 2020 Conference on Empirical Methods in
  Natural Language Processing: System Demonstrations}, pages 38--45, Online.
  Association for Computational Linguistics.

\bibitem[{Xiong et~al.(2019)Xiong, Huang, Guo, Zhou, Zhou, and
  Lin}]{xiong2019graph}
Yu~Xiong, Qingqiu Huang, Lingfeng Guo, Hang Zhou, Bolei Zhou, and Dahua Lin.
  2019.
\newblock \href {https://doi.org/10.1109/ICCV.2019.00469} {A graph-based
  framework to bridge movies and synopses}.
\newblock In \emph{2019 {IEEE/CVF} International Conference on Computer Vision,
  {ICCV} 2019, Seoul, Korea (South), October 27 - November 2, 2019}, pages
  4591--4600. {IEEE}.

\bibitem[{Yao et~al.(2020)Yao, Dai, Ramaswamy, Min, and Huang}]{yao2020weakly}
Wenlin Yao, Zeyu Dai, Maitreyi Ramaswamy, Bonan Min, and Ruihong Huang. 2020.
\newblock \href {https://doi.org/10.18653/v1/2020.emnlp-main.430} {Weakly
  {S}upervised {S}ubevent {K}nowledge {A}cquisition}.
\newblock In \emph{Proceedings of the 2020 Conference on Empirical Methods in
  Natural Language Processing (EMNLP)}, pages 5345--5356, Online. Association
  for Computational Linguistics.

\bibitem[{Yin et~al.(2021)Yin, Radev, and Xiong}]{yin-etal-2021-docnli}
Wenpeng Yin, Dragomir Radev, and Caiming Xiong. 2021.
\newblock \href {https://doi.org/10.18653/v1/2021.findings-acl.435}
  {{D}oc{NLI}: A large-scale dataset for document-level natural language
  inference}.
\newblock In \emph{Findings of the Association for Computational Linguistics:
  ACL-IJCNLP 2021}, pages 4913--4922, Online. Association for Computational
  Linguistics.

\bibitem[{Zhang et~al.(2020)Zhang, Zhao, Saleh, and Liu}]{zhang2019pegasus}
Jingqing Zhang, Yao Zhao, Mohammad Saleh, and Peter~J. Liu. 2020.
\newblock \href {http://proceedings.mlr.press/v119/zhang20ae.html} {{PEGASUS:}
  pre-training with extracted gap-sentences for abstractive summarization}.
\newblock In \emph{Proceedings of the 37th International Conference on Machine
  Learning, {ICML} 2020, 13-18 July 2020, Virtual Event}, volume 119 of
  \emph{Proceedings of Machine Learning Research}, pages 11328--11339. {PMLR}.

\bibitem[{Zhang et~al.(2021{\natexlab{a}})Zhang, Celikyilmaz, Gao, and
  Bansal}]{zhang2021emailsum}
Shiyue Zhang, Asli Celikyilmaz, Jianfeng Gao, and Mohit Bansal.
  2021{\natexlab{a}}.
\newblock \href {https://doi.org/10.18653/v1/2021.acl-long.537} {{E}mail{S}um:
  Abstractive email thread summarization}.
\newblock In \emph{Proceedings of the 59th Annual Meeting of the Association
  for Computational Linguistics and the 11th International Joint Conference on
  Natural Language Processing (Volume 1: Long Papers)}, pages 6895--6909,
  Online. Association for Computational Linguistics.

\bibitem[{Zhang et~al.(2021{\natexlab{b}})Zhang, Chen, and
  May}]{zhang2021salience}
Xiyang Zhang, Muhao Chen, and Jonathan May. 2021{\natexlab{b}}.
\newblock \href {https://doi.org/10.18653/v1/2021.emnlp-main.107}
  {Salience-aware event chain modeling for narrative understanding}.
\newblock In \emph{Proceedings of the 2021 Conference on Empirical Methods in
  Natural Language Processing}, pages 1418--1428, Online and Punta Cana,
  Dominican Republic. Association for Computational Linguistics.

\bibitem[{Zhao et~al.(2022{\natexlab{a}})Zhao, Huang, Chowdhury,
  Chandrasekaran, Mckeown, and Chaturvedi}]{zhao2022read}
Chao Zhao, Tenghao Huang, Somnath Basu~Roy Chowdhury, Muthu~Kumar
  Chandrasekaran, Kathleen Mckeown, and Snigdha Chaturvedi. 2022{\natexlab{a}}.
\newblock Read top news first: A document reordering approach for
  multi-document news summarization.
\newblock In \emph{Findings of the Association for Computational Linguistics:
  ACL 2022}, pages 613--621.

\bibitem[{Zhao et~al.(2022{\natexlab{b}})Zhao, Yao, Yu, Song, Yu, and
  Chen}]{zhao2022learning}
Chao Zhao, Wenlin Yao, Dian Yu, Kaiqiang Song, Dong Yu, and Jianshu Chen.
  2022{\natexlab{b}}.
\newblock Learning-by-narrating: Narrative pre-training for zero-shot dialogue
  comprehension.
\newblock In \emph{Proceedings of the 60th Annual Meeting of the Association
  for Computational Linguistics (Volume 2: Short Papers)}, pages 212--218.

\bibitem[{Zhong et~al.(2019)Zhong, Liu, Wang, Qiu, and
  Huang}]{zhong-etal-2019-searching}
Ming Zhong, Pengfei Liu, Danqing Wang, Xipeng Qiu, and Xuanjing Huang. 2019.
\newblock \href {https://doi.org/10.18653/v1/P19-1100} {Searching for effective
  neural extractive summarization: What works and what{'}s next}.
\newblock In \emph{Proceedings of the 57th Annual Meeting of the Association
  for Computational Linguistics}, pages 1049--1058, Florence, Italy.
  Association for Computational Linguistics.

\bibitem[{Zhong et~al.(2021)Zhong, Yin, Yu, Zaidi, Mutuma, Jha, Awadallah,
  Celikyilmaz, Liu, Qiu, and Radev}]{zhong2021qmsum}
Ming Zhong, Da~Yin, Tao Yu, Ahmad Zaidi, Mutethia Mutuma, Rahul Jha,
  Ahmed~Hassan Awadallah, Asli Celikyilmaz, Yang Liu, Xipeng Qiu, and Dragomir
  Radev. 2021.
\newblock \href {https://doi.org/10.18653/v1/2021.naacl-main.472} {{QMS}um: A
  new benchmark for query-based multi-domain meeting summarization}.
\newblock In \emph{Proceedings of the 2021 Conference of the North American
  Chapter of the Association for Computational Linguistics: Human Language
  Technologies}, pages 5905--5921, Online. Association for Computational
  Linguistics.

\end{thebibliography}
\bibliographystyle{acl_natbib}

\appendix

\section{Appendix}
\label{sec:appendix}

\subsection{Compression-aware Evaluation}
\label{app::split}

\begin{table}[t!]
		\centering
		\small
		\setlength{\tabcolsep}{0.4em} %
		\begin{tabular}{lccccc}
\hline \textbf{Subsets}  & \textbf{Size}  & \textbf{L-doc} & \textbf{L-sum} & \textbf{Factor} & \textbf{Ratio} \\\hline
\rowcolor{Gray!93}\textit{\textbf{Validation}}& & & & & \\
\hline Low Comp. & 1,837  & 461 & 170 & 0.37 & 2.70 \\
\hline Medium Comp. & 2,161  & 704 & 152 & 0.22 & 4.55 \\
\hline High Comp. & 1,736  & 1,290 & 103 & 0.09 & 11.11 \\
\hline\rowcolor{Gray!93}\textit{\textbf{Test}}& & & & &\\
\hline Low Comp. & 1,773  & 443 & 164 & 0.37 & 2.70 \\
\hline Medium Comp. & 2,160 & 696 & 151 & 0.22 & 4.55 \\
\hline High Comp. & 1,590  & 1,355 & 108 & 0.09 & 11.11 \\
\hline
\end{tabular}

		\caption{\label{tab::stat_split} Statistics of validation and test subsets. The compression factor (Factor) is defined as the length ratio between the summary to the document. The compression ratio (Ratio) is defined as the length ratio between the document to the summary.} %
\end{table}

\begin{table*}[t]

		\small
		\setlength{\tabcolsep}{0.47em} %
	\begin{center}
\renewcommand{\arraystretch}{1.2}
\begin{tabular}{l cccccccccccc}
\toprule
    \multirow{3}*{Model} & \multicolumn{4}{c}{Low Comp (Long Summ.)} & \multicolumn{4}{c}{Medium Comp. (Medium Summ.)}  & \multicolumn{4}{c}{High Comp. (Short Summ.)}  \\
    \cmidrule(lr){2-5}\cmidrule(lr){6-9}\cmidrule(lr){10-13}
    & \textbf{\textsc{R-1}} & \textbf{\textsc{R-2}} & \textbf{\textsc{R-L}} &\textbf{\textsc{SC}} &
    \textbf{\textsc{R-1}} &\textbf{\textsc{R-2}} & \textbf{\textsc{R-L}} & \textbf{\textsc{SC}} &
    \textbf{\textsc{R-1}} & \textbf{\textsc{R-2}} & \textbf{\textsc{R-L}} & \textbf{\textsc{SC}} \\
\midrule
\rowcolor{Gray!93}\textit{\textbf{Extractive}}& & & &&&&&&&&&\\
            \textsc{Rand} & 37.19 & 7.14 & 32.64 & - & 34.79 & 5.35 & 30.72 & - & 29.16 & 3.43 & 25.38 & - \\
\textsc{Lead} & 38.62 & 8.45 & 33.88 &  - & 36.18 & 6.87 & 31.92 &  - & 29.73 & 4.54 & 25.9 &  - \\
\textsc{LexRank} & 36.90 & 7.14 & 32.11 &  - & 35.11 & 5.82 & 30.68 &  - & 30.02 & 4.17 & 25.66 &  - \\
\textsc{TextRank} & 37.55 & 7.67 & 32.77 &  - & 35.89 & 6.23 & 31.22 &  - & 30.77 & 4.43 & 26.21 &  - \\ \hline
\textsc{HSG} & 39.86 & 9.06 & 34.97 &  - & 38.00 & 7.74 & 33.47 &  - & 32.25 & 5.54 & 27.90 &  - \\
\textsc{Bert-B} & 39.00 & 8.82 & 34.04 &  - & 37.51 & 7.50 & 32.91 &  - & 31.77 & 5.29 & 27.49 &  - \\
\textsc{Roberta-B} & 39.08 & 8.78 & 34.06 &  - & 37.60 & 7.50 & 32.96 &  - & 32.01 & 5.40 & 27.69 &  - \\
\textsc{LFormer-B} & \textbf{40.38}* & \textbf{9.41}* & \textbf{35.26} &  - & \textbf{38.69}* & \textbf{8.08}* & \textbf{33.88}* & - & \textbf{32.99}* & \textbf{5.85}* & \textbf{28.27}* &  - \\\hline
\textsc{Oracle} & 43.55 & 12.48 & 37.89 &  - & 43.36 & 11.6 & 37.72 &  - & 39.88 & 10.06 & 33.81 &  - \\
			\midrule
\rowcolor{Gray!93}\textit{\textbf{Abstractive}}& & & &&&&&&&&&\\
\textsc{Bart-B} & 38.21 & 8.80 & 33.8 &  71.94 & 36.62 & 7.51 & 32.60 &  66.85 & 32.02 & 5.99 & 28.21 &  57.14  \\
\textsc{T5-B} & 38.56 & 8.74 & 34.14 & 81.43 & 37.34 & 7.60 & 33.23 &  78.07 & 32.61 & 5.68 & 28.53 &  \textbf{69.81}* \\
\textsc{LED-B} & 39.46 & 9.22 & 34.93 &  74.01 & 38.10 & 8.18 & 33.96 &  66.85 & 33.58 & 6.83 & 29.53 &  47.34  \\\hline
\textsc{Bart-L} & 39.29 & 9.37 & 34.87 &  \textbf{86.86}* & 36.92 & 8.04 & 32.94 &  \textbf{81.48}* & 33.84 & 7.10 & 29.67 &  64.05\\
\textsc{T5-L} & 39.39 & 9.17 & 34.94 & 84.09 &  \textbf{38.49}* & 8.22 & \textbf{34.29}* & 75.65 & 34.65 & 6.79 & 30.46 &  63.43\\
\textsc{Pegasus-L} & 39.21 & 9.14 &   34.67 & 84.70 &  37.57 & 7.86 & 33.34 &  80.70 &  33.65 & 6.67 & 29.40 &  60.13 \\
\textsc{LED-L} & \textbf{39.57} & \textbf{9.74}* & \textbf{35.06} &  78.78  & 37.86 & \textbf{8.73}* & 33.65 &  71.43 &  \textbf{35.41}* & \textbf{8.11}* & \textbf{30.99}* &  50.83 \\
 \bottomrule[1pt]

		\end{tabular}
		
	\end{center}
	\caption{\label{tab::results_comp} Summarization results evaluated on three test subsets of \textsc{NarraSum} over ROUGE 1 (R-1), ROUGE 2 (R-2), ROUGE L (R-L), and SummaC (SC). We highlight the best scores separately for extractive and abstrative systems. * indicates a statistically significant difference compared with the second best score (bootstrap resampling, $p<0.05$ \cite{koehn2006manual}). }
	\vspace{-1em}
\end{table*}

Intuitively, creating a short summary is more challenging than creating a long one, as it requires selecting information more precisely and textualizing the salient information more abstractively (for abstractive models only). 

To better understand the models' capability under different degrees of compression, we split the validation set and the test set into three subsets based on the compression factor, $\alpha$, of the gold summary. The compression factor is defined as the length ratio between the summary to document.
Specifically, we regard $\alpha < 0.15$ as \textbf{high compression}, $0.15 \leq \alpha < 0.3$ as \textbf{medium compression}, and $\alpha \geq 0.3$ as \textbf{low compression}. Using these threshold values, we can split the validation and test sets into three similar-sized subsets. We list the detailed statistics of each subset in Table \ref{tab::stat_split}. 

During inference, the desired length of the summary for a given document is determined by multiplying the document's length (the number of tokens in the document) with the $\alpha$ for the desired compression level. The $\alpha$s for the three compression levels are determined using their average values in the corresponding validation subsets (0.37, 0.22, and 0.09 for the low, medium, and high compression, respectively). 
For extractive models, we continually add sentences to the summary according to their predicted salience until the summary length is most close to the desired length (either longer or shorter). For the abstractive models, we roughly control the length of the summary
using the method described by \citet{fan2018controllable}. The basic idea is to split the entire dataset into $n$ equally-sized bins according to the summary length. For each data instance, the ID of the corresponding bin is appended to the front of the document to indicate the desired scope of summary length. We use $n=10$.

Table \ref{tab::results_comp} shows the results on \textsc{NarraSum} using extractive and abstractive summarization approaches for different degrees of compression. Generally, with the increase in the compression ratio, the Rouge scores become lower. It indicates that it is more challenging to create a short summary compared with a long one. However, observations on the entire test set still hold across different levels of compress degree.

\begin{table}[h!]
    \centering
    \small
    \renewcommand{\arraystretch}{2.3}
    \begin{tabular}{p{0.99\linewidth}}
    \toprule
         \textbf{Document:} It is the girls' second year at PCA. Dana has been accepted to a European Exchange Program, so Zoey and Nicole think will have their dorm room to themselves. However, Coco informs them they will be getting a new roommate. Zoey and Nicole go to the housing office hoping to choose a roommate, but their request is denied (After Nicole accidentally destroy's the secretary's perfume collection). When they return to their dorm, they find their new roommate, Lola Martinez, a goth-punk girl who disturbs them by drinking raw eggs and claiming to talk to the dead. As Lola's behavior grows more erratic, Zoey and Nicole are out of the dorm, but they later discover that Lola is not a goth-punk, but an aspiring actress who managed to fool them with her acting skills. Meanwhile, Michael and Chase have to deal with Logan, whose father has given him a huge entertainment center for their dorm room—which turns out to be a huge headache. \\\hline
         
         \textbf{Gold Summary:} Zoey and Nicole return to PCA for the second year. But they soon find out Dana will not be returning, and a very strange roommate, Lola, has come to take her place. The year gets to a bad start for Chase and Michael, because of Logan's very loud gifts from his dad.  \\\midrule
         
         \textbf{Bart-large:} \textbf{\textcolor{violet}{[Missing \textit{Zoey returns to PAC} and \textit{DANA will not return}]}} Zoey and Nicole get a new roommate, Lola Martinez, a goth-punk girl who disturbs them by drinking raw eggs and claiming to talk to the dead. Meanwhile, Michael and Chase have to deal with Logan, whose father has given him a huge entertainment center for their dorm room--which turns out to be a huge headache.\\\hline
         \textbf{T5-Large:} Zoey and Nicole get a new roommate, Lola Martinez, who turns out to be a goth-punk. Meanwhile, Michael and Chase have to deal with Logan, whose father has given him a huge entertainment center for their dorm room. \textbf{\textcolor{violet}{[Fail to follow the causal and temporal relationship]}} Meanwhile, Dana is accepted to a European exchange program, so the girls think they'll have their own room, but Coco tells them that they'll be getting a roommate.\\\hline
         \textbf{Pegasus-Large:} Zoey and Nicole's first year at PCA is coming to an end, and they're looking forward to living in their own dorm room. However, Coco informs them that Dana has been accepted to a European exchange program, so Zoey and Nicole think \textbf{\textcolor{violet}{[Not logically coherent]}} they will have their own room to themselves. Zoey and Nicole go to the housing office hoping to choose a roommate, but their request is denied (After Nicole accidentally destroy's the secretary's perfume collection).\\\hline
         \textbf{LED-Large:} Zoey and Nicole are thrilled when Dana gets accepted to a European exchange program at PCA, \textbf{\textcolor{violet}{[Not fluent]}} but when they find out they have to share a dorm room with Lola Martinez, a goth-punk girl who drinks raw eggs and talks to the dead. Meanwhile, Michael and Chase have to deal with Logan, whose father has given him a huge entertainment center for their dorm room, which turns out to be a huge headache. \\\bottomrule
    \end{tabular}
    \caption{Sample summaries generated by baseline models. We show the original document, the gold summary, and the output summaries from four large models. We highlight the typical errors of each output summary. }
    \label{tab::case}
\end{table}

\subsection{Qualitative Analysis}
\label{app::case}
Table \ref{tab::case} shows an example with the narrative document, gold summary, and the predicted summaries. The narrative document is from Season 2, Episode 1 of \textit{Zoey 101}, an American comedy-drama TV. 

This example shows that while the gold summary can faithfully cover the most salient information from the narrative document, summaries generated by machines contain some errors. Bart does not contain the information of ``\textit{Zoey returns to PAC}'' and ``\textit{Dana will not return}''.
T5 fails to follow the causal and temporal relationships of events. The summary created by Pegasus is generally not coherent. The summary created by LED covers all important information but the writing is not fluent.

\subsection{Zero-Shot Tasks}
\label{app::zero-shot}
We choose five tasks that are designed for narrative understanding (MCTest, MovieQA,
LiSCU, CBT, and QuAIL), and one task for narrative summarization (Reddit TIFU).
We don't include tasks that are not related to narrative. 
In this section, we describe the details of these datasets.%

MCTest \cite{richardson2013mctest} is a dataset designed for open-domain multiple-choice reading comprehension. The dataset contains 500 fictional stories, with four multiple choice questions per story. 

CBT \cite{hill2016goldilocks} is also an dataset designed for open-domain reading comprehension. The dataset builds question-answer pairs from 108 children’s books with clear narrative structure.

MovieQA \cite{tapaswi2016movieqa} aims to evaluate models' ability of automatic story comprehension. The dataset consists of 14,944  multiple-choice questions sourced from  408 movies. Each question has five options. We use the movie summaries as input to answer these questions. 

LiSCU \cite{brahman2021let} is a character-centric narrative understanding task to test the model performance from the perspective of characters.
This dataset contains 1,708 literature summaries and 9,499 character descriptions.
Given the literature summary, the model needs to identify the character’s name from an anonymized character description and a list of character candidates.

QuAIL \cite{rogers2020getting} is a machine reading comprehension benchmark with varying types of reasoning. Solving this challenge requires an understanding of not only the text-based information from the document but also the world knowledge and commonsense knowledge. Documents in  QuAIL are collected from fiction, user stories, and so on. Each question has four options.

Reddit TIFU \cite{kim2019abstractive} is an abstractive summarization dataset. It consists of 120K crowd-generated posts from the online discussion forum Reddit, as well as their corresponding summaries. Different from other narrative summarization datasets we discussed in the paper, narratives in Reddit TIFU are mostly written in informal and conversational text, and the story is about the poster doing something wrong or messing everything up. These features make Reddit TIFU a good out-of-domain test data to evaluate the models' generalization power for narrative summarization.

\end{document}